\pdfoutput=1

\documentclass{article}
\usepackage{iclr2027_preprint,times}

\usepackage{amsmath,amsfonts,bm}

\def\eqref#1{equation~\ref{#1}}

\def\1{\bm{1}}

\DeclareMathAlphabet{\mathsfit}{\encodingdefault}{\sfdefault}{m}{sl}
\SetMathAlphabet{\mathsfit}{bold}{\encodingdefault}{\sfdefault}{bx}{n}

\usepackage{hyperref}
\usepackage{url}
\usepackage{amsmath,amssymb,booktabs,graphicx,xcolor,multirow,adjustbox,longtable}

\title{One Evaluation, Any Operating Point: Hypernetwork-Amortized MeanFlow for 3D MRI Reconstruction}

\author{Ruibo Wang\thanks{Ruibo Wang is with the Faculty of Electrical Engineering, Mathematics and Computer Science, Delft University of Technology, 2600 AA Delft, Netherlands. Correspondence: \texttt{ruibowang953@gmail.com}.} \\
Delft University of Technology \\
}

\iclrfinalcopy
\begin{document}
\maketitle

\begin{abstract}
Generative priors reconstruct accelerated 3D MRI well but pay twice at deployment: tens of network evaluations
per volume, and hyperparameters re-tuned per protocol. A third cost is never counted because it is fixed before
the model exists: the sampling pattern the scanner uses. We treat the whole operating point as an input. A 3D
MeanFlow patch network (a one-step flow model, first trained as an average-velocity prior on volumes alone) is
fine-tuned end-to-end from the zero-filled image through a warm-started, five-iteration differentiable
conjugate-gradient projection, and a small hypernetwork maps the operating point (the data-consistency weight, the acceleration, and the Cartesian sampling pattern
itself) to the network's
per-channel modulation. Three findings follow. (i) Learning the acquisition is worth far more than any other
operating point: on fully sampled clinical knee acquisitions with scanner-derived coil maps, the learned mask
gains $+2.34/+1.57/+0.86/+0.67$~dB at $4/8/16/32\times$ over the protocol's variable-density mask, on $10/10$
volumes at every rate, and the gain survives a stronger reconstructor (it \emph{grows} at three evaluations). The gain needs the
solver: with a LOUPE-style feed-forward reconstructor the same learned mask \emph{hurts} at $4\times$
($-1.7$~dB), with the data-consistency projection it is worth $+4.6$~dB on $150/150$ volumes. (ii) One evaluation is a real
operating point: it beats a 20-step patch-diffusion prior re-trained on the same data by $+1.5/+3.1/+3.1$~dB on
brain and $+0.8/+2.0/+2.9$~dB on knee, and three and five evaluations extend the front to $+6$~dB while using a
quarter of the prior's network calls. (iii) The fully sampled archive is optional: trained with no target, on a
split of its own acquired samples, the reconstructor matches its supervised twin at $4\times$ on real data
($39.93$ vs.\ $39.92$~dB). We also report, with the measurements that explain them, what did not work:
subject-adaptive acquisition from measured energy, self-supervision combined with learned acquisition, and
amortising the data-consistency weight or a loss trade-off.
\end{abstract}

\section{Introduction}
Generative priors have become the strongest class of reconstructors for accelerated 3D MRI: a patch-based
diffusion prior trained on volumes alone, combined with conjugate-gradient (CG) data consistency at inference,
reconstructs multi-coil knee and brain data at $4$ to $16\times$ with state-of-the-art fidelity and needs no paired
training data \citep{hu2024learning,chung2022score,jalal2021robust}. Two costs are paid at deployment. The first is time:
every reconstruction runs the network tens of times, each step followed by a CG solve, and a clinical-size 3D
volume takes minutes on a workstation GPU. The second is tuning: the data-consistency weight $\lambda$, the
acceleration the prior was tuned for and the sampler's schedule are hyperparameters of the
\emph{reconstruction}, not of the prior; they are re-tuned per protocol and, because no reference image exists
at scan time, cannot be tuned per patient.

There is a third cost, usually not counted as one because it is fixed before the model exists: the
\emph{acquisition}. Which phase-encode lines a scanner collects is set by the protocol, and every reconstructor
in this literature is trained to undo that particular choice. This paper's main finding is that the choice is
the most valuable thing to hand to the model: on clinical acquisitions, letting the network pick the Cartesian
sampling pattern for a requested acceleration is worth two to four times more than any reconstruction
hyperparameter we amortise, and, unlike those, its value grows as the reconstructor improves.

\begin{figure}[t]
\centering
\includegraphics[width=0.74\textwidth]{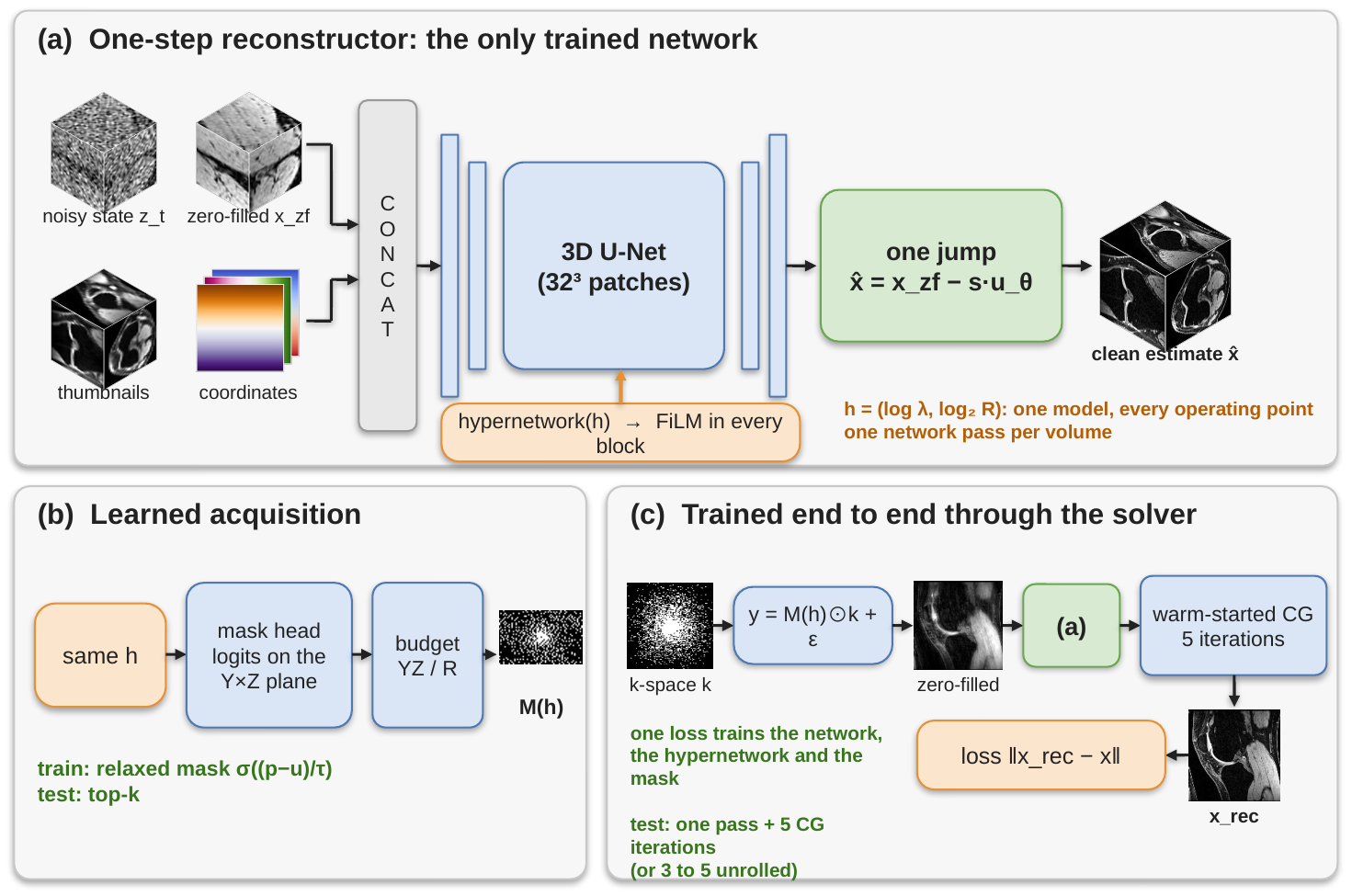}
\caption{\textbf{(a)} The measurement-conditioned one-step reconstructor: a $32^3$ patch of the noisy state, the
zero-filled condition, pooled thumbnails of both and the voxel coordinates enter a plain 3D U-Net whose blocks
are modulated (FiLM) by a hypernetwork of the operating point $h=(\log\lambda,\log_2R)$; one jump
$\hat x = x_{\mathrm{zf}} - s\,u_\theta$ gives the clean estimate. \textbf{(b)} The same $h$ drives a second
hypernetwork head that emits logits over the phase-encode plane, renormalised to the budget $1/R$: a relaxed mask
in training, top-$k$ at test time. \textbf{(c)} Training runs the whole chain (mask, measurement, zero-filled image, one evaluation, warm-started five-iteration
CG) and the loss gradient reaches the hypernetwork and the
mask logits through the solver. At test time: one evaluation plus CG, or three to five unrolled evaluations.}
\label{fig:arch}
\end{figure}

One-step generative models \citep{song2023consistency,geng2026mean} and amortised hyperparameter learning
\citep{hoopes2021hypermorph,wang2022computing,ha2016hypernetworks} answer the first two costs in adjacent problems, and LOUPE
\citep{bahadir2019learning,bahadir2020deep} learns a sampling pattern jointly with a feed-forward 2D reconstructor. Combining these is
not a matter of swapping the loss: we show by measurement that a one-step \emph{prior} used inside plug-and-play
falls below the prior-free baseline, that noising the network input destroys the measurement, that the MeanFlow
jump vanishes in the deterministic limit, and that a learned acquisition attached to a feed-forward reconstructor
is worth nothing or less. The design that works (Fig.~\ref{fig:arch}) is a \emph{measurement-conditioned} MeanFlow
reconstructor: the MeanFlow patch network with the zero-filled image as a condition, the one-step prior as its
initialisation, and end-to-end training through a warm-started five-iteration differentiable CG projection, with
a hypernetwork over the operating point $(\lambda, R)$ that also emits the sampling pattern for $R$. To our
knowledge it is the first one-step flow (MeanFlow) reconstructor for 3D multi-coil MRI; we are explicit below about
what the flow parametrisation buys and what it does not.

We evaluate under one fixed protocol for every arm (the same evaluation volumes, forward operator, noise level,
metrics and $\lambda$ grid) and keep a 20-step patch-diffusion prior re-trained in our pipeline as the
multi-step reference, so that every comparison isolates the method.

\paragraph{Contributions.}
\begin{enumerate}\setlength{\itemsep}{1pt}
  \item \textbf{The acquisition as an operating point, on real data, and when it pays.} A hypernetwork emits the
  3D Cartesian sampling pattern for a requested acceleration and is trained jointly with a multi-coil
  reconstructor through the data-consistency solver. On SKM-TEA knee data this gains
  $+2.34/+1.57/+0.86/+0.67$~dB at $4/8/16/32\times$ over the protocol's variable-density mask ($10/10$ volumes
  at every rate), survives four seeds, three hand-tuned Gaussian baselines and a second machine, grows with a three-evaluation reconstructor and persists
  at five, and is orthogonal to the hypernetwork. A controlled $2\times2$ shows that the
  gain \emph{requires} enforced data consistency: the same learned mask is $-1.7$~dB with a LOUPE-style
  feed-forward reconstructor and $+4.6$~dB with the projection, on all 150 validation volumes (Sec.~\ref{sec:acqexp}).
  \item \textbf{A MeanFlow reconstructor whose cost-quality front dominates multi-step priors.} One measurement-conditioned evaluation
  is $+1.5/+3.1/+3.1$~dB above a 20-step patch-diffusion prior on brain ($60^3$; $+1.4/+3.4/+4.5$ at $120^3$) and
  $+0.8/+2.0/+2.9$~dB on knee; five evaluations reach $+6.2/+5.9/+4.6$~dB with four times fewer network calls
  (Sec.~\ref{sec:front}).
  \item \textbf{An amortised operating point, and which hyperparameters deserve it.} One FiLM hypernetwork covers
  $\lambda\in[10^{-3},0.3]$ and $R\in[4,32]$, matches or beats every single-point specialist, interpolates to
  unseen accelerations and transfers zero-shot to twice the resolution. The acceleration is worth amortising;
  the data-consistency weight and a fidelity/sharpness loss trade-off are not (Sec.~\ref{sec:amort}).
  \item \textbf{No fully sampled archive required, and negative results with their mechanisms.} Trained only on a
  split of its own acquired samples, the reconstructor matches its supervised twin at $4\times$ on real data.
  Subject-adaptive acquisition from measured energy, self-supervision combined with learned acquisition, and
  stochastic one-step sampling all fail, each for a reason we measure (Secs.~\ref{sec:ssl} and \ref{sec:neg}).
\end{enumerate}

\section{Background and related work}\label{sec:bg}
\paragraph{Generative priors and plug-and-play reconstruction.} Score- and diffusion-based priors reconstruct
undersampled MRI by alternating a learned denoising step with a data-consistency step
\citep{jalal2021robust,chung2022score,song2021solving,chung2022diffusion}; patch-based diffusion priors \citep{wang2023patch}
make this affordable at 3D volume resolution by training on $32^3$ patches conditioned on a pooled thumbnail of
the whole volume and on the patch position, as \citet{hu2024learning} do for inverse problems. Our multi-step
reference is our own 3D MRI re-implementation of that recipe, an EDM denoiser \citep{karras2022elucidating} on $32^3$ patches, tiled over the volume, alternated
with an early-stopped CG solve of $(A^{\!H}\!A+\lambda I)z=A^{\!H}y+\lambda\hat x$ from $z_0=0$, re-trained on our data with our data-consistency code and metrics.
\paragraph{One-step generation.} Consistency models \citep{song2023consistency} and average-velocity objectives
\citep{geng2026mean} learn $u(z_t,r,t)=\frac{1}{t-r}\int_r^t v\,d\tau$ on the rectified-flow path
\citep{liu2022flow,lipman2022flow}, so that $z_r=z_t-(t-r)\,u$ is exact for any interval and one network
evaluation is a sampler. We use a MeanFlow patch prior as the \emph{initialisation} of our reconstructor and show
that on its own it is a poor plug-and-play denoiser.
\paragraph{Unrolled reconstruction.} MoDL \citep{aggarwal2018modl}, the variational network \citep{hammernik2018learning,sriram2020end} and
their descendants alternate a learned block with a data-consistency solve for a fixed number of iterations,
trained end-to-end on paired data. Our reconstructor is the one-iteration member of that family with a
generative initialisation and an amortised operating point; we train unrolled baselines with the same backbone,
data and losses so the comparison isolates those differences.
\paragraph{Amortised hyperparameters.} HyperMorph \citep{hoopes2021hypermorph} trains a registration network over a
distribution of regularisation weights with a hypernetwork producing the primary network's weights; HyperRecon
\citep{wang2022computing} carries the idea to unrolled MRI reconstruction, amortising the data-consistency weight.
Generating all weights of a 45M-parameter 3D network is not viable; we use per-block FiLM modulation
\citep{perez2018film} and compare it with rank-$k$ kernel deltas. Implicit-kernel convolutions \citep{ma2023hyper} are a
related idea at the level of the kernel.
\paragraph{Learned and active acquisition.} LOUPE \citep{bahadir2019learning,bahadir2020deep} learns a probabilistic Cartesian mask
jointly with a feed-forward reconstructor through a sigmoid relaxation; earlier work chose patterns greedily
\citep{gozcu2018learning}, and later work learns non-Cartesian trajectories \citep{weiss2019pilot} or acquires actively per subject
\citep{zhang2019reducing,pineda2020active}. We use LOUPE's relaxation inside the hypernetwork, in 3D and multi-coil,
conditioned on the acceleration, and trained through the data-consistency solver; we measure that the last point is what makes it pay. Self-supervised training without fully sampled data follows SSDU \citep{yaman2020self}.

\section{Method}\label{sec:method}
\paragraph{Setting.} A volume $x\in\mathbb C^{N}$ is observed through $S$ coil sensitivities as
$y = M\mathcal F S x + \varepsilon =: Ax+\varepsilon$, with $M$ a Cartesian mask on the phase-encode plane
sampling a fraction $1/R$ of the lines and $\varepsilon$ complex Gaussian noise. The zero-filled image is
$x_{\mathrm{zf}}=A^{\!H}y$; CG-SENSE \citep{pruessmann1999sense,pruessmann2001advances} solves $\min_z\|Az-y\|^2$ by early-stopped CG and is our
prior-free control. Every method below shares $A$, the CG code, the noise realisations and the metrics.

\subsection{A measurement-conditioned one-step reconstructor}\label{sec:onestep}
The backbone is a plain 3D U-Net $u_\theta$ on $32^3$ patches (45M parameters), with the patch-plus-thumbnail
input that patch-based diffusion priors use \citep{wang2023patch,hu2024learning}: its input is a patch of the state, the whole state average-pooled to $32^3$, and the
voxel coordinates, and a volume is processed by tiling. We first train it as a MeanFlow prior
\citep{geng2026mean} on volumes alone (App.~\ref{app:prior}); this is only the initialisation. The reconstructor
takes the zero-filled image both as its \emph{starting point} and as a \emph{condition}: with
$z=(1-t_s)x_{\mathrm{zf}}+t_s\epsilon$, $\epsilon\sim\mathcal N(0,I)$, the input is $[z, x_{\mathrm{zf}}]$
(patch and thumbnail of each, 11 channels with the coordinates; the prior's stem is extended with zero weights
so that initialisation reproduces the prior exactly), and one jump gives the clean estimate
\begin{equation}
\hat x \;=\; x_{\mathrm{zf}} - s\,u_\theta\big([z, x_{\mathrm{zf}}], t_s, h\big),\qquad s=\max(t_s,\tfrac12).
\label{eq:jump}
\end{equation}
The residual is taken from the \emph{clean} condition rather than from $z$: the MeanFlow form $\hat x=z-t_s u$
is an $\epsilon$-prediction problem at large $t_s$ and makes the network vanish in the deterministic limit
$t_s\to0$ (the two coincide at $t_s=0$). A warm-started differentiable CG projection then enforces the data,
\begin{equation}
x_{\mathrm{dc}} \;=\; \mathrm{CG}_5\!\Big((A^{\!H}\!A+\lambda I)\,z = A^{\!H}y+\lambda\hat x;\; z_0=\hat x\Big),
\label{eq:cg}
\end{equation}
and the loss is $\|x_{\mathrm{dc}}-x\|^2+\|\hat x-x\|^2$ with $\lambda\sim\log\mathcal U[10^{-3},1]$,
$R\sim\{4,8,16,32\}$ re-masked on the fly, and $t_s\sim\mathcal U[0.05,0.7]$ (the deterministic model uses
$t_s=0$ at test time). \emph{Why warm-started:} a cold-started truncated CG ($z_0=0$) barely propagates $\hat x$ to the unsampled
frequencies in five iterations, so the loss is flat and the network stays at its initialisation (App.~\ref{app:design});
started at $\hat x$, the same five iterations are a light projection of the network's estimate onto the data, and
the gradient reaches the network. At test time one evaluation of \eqref{eq:jump} and one projection \eqref{eq:cg} is the reconstruction;
repeating the pair $K$ times, each iteration starting from the current estimate and still conditioned on
$x_{\mathrm{zf}}$, gives the unrolled (MoDL/VarNet-form) members of the same family that we use as baselines and
as the strong end of the cost-quality front.

\subsection{The operating point as an input}\label{sec:hyper}
The operating point $h=(\log\lambda,\log_2R)$ (and $t_s$ for the stochastic variant) is normalised to
$[-1,1]^2$ and mapped by a three-layer MLP to an embedding that sets the FiLM scale and shift after every
normalisation of every U-Net block (0.18M hypernetwork parameters). HyperMorph generates the primary network's weights outright, which is not viable at 45M parameters; we compare
FiLM with rank-4 additive deltas on every $3^3$ kernel (1.2M parameters) and FiLM is $0.3$ to $0.4$~dB better at equal
budget (App.~\ref{app:extended}): the modulation needed here is a per-channel gain, not a change of filter shape. A plain MLP on the
normalised inputs interpolates between training rates; a Fourier-feature lift, our first choice, does not
(Sec.~\ref{sec:amort}). At test time $R$ is known from the mask and $\lambda$ can be chosen per subject without a
reference by the discrepancy principle \citep{morozov1966solution} on the sampled-k-space residual; in practice the amortised
model is flat in $\lambda$ to $0.03$~dB and a default is within that of the oracle (App.~\ref{app:extended}).

\subsection{The acquisition as part of the operating point}\label{sec:acq}
A second head of the same hypernetwork maps $h$ to logits $\ell(h)\in\mathbb R^{Y\times Z}$ over the
phase-encode plane. Probabilities $p=\sigma(\ell)$ are renormalised so that $\sum p = YZ/R$ (rescaling below one
or its complement above one, as in LOUPE), and a mask is drawn by the relaxation
$M=\sigma\big((p-u)/\tau\big)$, $u\sim\mathcal U[0,1]$, during training and by keeping the top-$YZ/R$ probabilities
at test time. The measurement $y=M\odot\mathcal F S x+\varepsilon$ is formed from stored fully sampled k-space,
so the gradient of the reconstruction loss flows through the CG projection, the network and the relaxed mask into
$\ell(h)$; the mask parameters get a $10\times$ learning rate. The logits live on a fixed grid and are resampled
bilinearly to the volume's phase-encode plane, so one policy serves $60^3$ and $120^3$ brain volumes and the
$128\times128\times80$ knee alike. 

\subsection{Without fully sampled targets}\label{sec:ssl-method}
Following SSDU \citep{yaman2020self}, the acquired lines are split into a visible set (60\%, always containing the
k-space centre) and a held-out set; the model reconstructs from the visible set and is scored on how well
$A_{\mathrm{held}}\,x_{\mathrm{dc}}$ predicts the held-out samples. Nothing else changes: the same network,
projection and hypernetwork, and the same learned-acquisition head can be attached, which is where Sec.~\ref{sec:ssl} finds a failure worth reporting.

\section{Experiments}\label{sec:results}
\paragraph{Protocol.} PSNR on the magnitude over the whole volume (target maximum as range) and per-slice SSIM,
best $\lambda$ of a common grid for every method, 10 evaluation volumes unless a table says all 150, paired
per-volume comparisons with 95\% intervals. Data: SKM-TEA knee \citep{desai2022skm} at quarter resolution ($128{\times}128{\times}80$) from the
fully sampled raw acquisitions with scanner-derived coil maps (\emph{real} data, real sensitivities); and
BraTS \citep{baid2021rsna,de20242024} at $60^3$ and $120^3$ with eight simulated coils. Undersampling is a variable-density
Gaussian mask on the phase-encode plane with a fully sampled centre (the ``protocol mask''); complex Gaussian
noise is added at 30~dB measured on the sampled points. Two noise conventions differ by $1$ to $2$~dB and synthesising $y$ from the image inflates every iterative
reconstructor by up to $2.4$~dB at $4\times$; both are measured in App.~\ref{app:protocol}, with a reproducibility test. The BraTS $60^3$ results come from two splits of the same data, a stored-measurement split (the front, Sec.~\ref{sec:front}) and a
full-k-space split (learned acquisition, Sec.~\ref{sec:acqexp}); each table names its split.

\subsection{Learning the acquisition}\label{sec:acqexp}
Table~\ref{tab:acq} compares the learned mask against an identical model trained and evaluated with the
protocol mask on three datasets. On the \emph{real} knee acquisitions the learned pattern wins at every rate and
on every volume, $+2.34/+1.57/+0.86/+0.67$~dB at $4/8/16/32\times$; at 10k instead of 30k steps the same pair
reads $+2.06/+1.19/+0.81/+0.60$, so training both sides longer widens the gap; on all 33 validation volumes that
10k-step pair gains $+2.05/+1.26/+0.92/+0.72$~dB, $33/33$ at every rate. On simulated brain volumes the
gain is larger where the budget is comfortable ($+4.64$~dB at $4\times$, $60^3$, $150/150$ volumes) and reverses at $16$ to $32\times$
on the small $60^3$ volumes, but not on $120^3$ where each volume carries eight times more data. The reversal is
a budget effect, not a rate-sharing effect (10-volume controls): a sampler trained for one rate is no better than
the shared one ($44.22$ vs.\ $44.56$~dB at $4\times$), and one trained only at $16\times$ still loses $0.97$~dB to
the protocol mask. Fig.~\ref{fig:masks_app} (App.~\ref{app:extended}) shows why: the learned pattern spends its budget on the centre (fraction of the
central $12\times12$ block kept: $0.69/0.59/0.50/0.44$ vs.\ $0.59/0.38/0.20/0.07$), the right trade until only a
few percent of the plane can be sampled. The comparison is robust: a second seed of the control agrees to $0.03$~dB, four seeds across two machines and
data paths reproduce the pair within $0.02$ to $0.35$~dB, and no hand-tuned Gaussian reaches it; a narrower mask, the hand-made way to ``concentrate on the centre'', is
the worst of three widths (App.~\ref{app:extended}). What the learned pattern buys is its shape.

\begin{table}[t]\centering\footnotesize\setlength{\tabcolsep}{4pt}
\caption{\textbf{Learning the acquisition.} PSNR (dB) of the same reconstructor trained and evaluated with the
protocol variable-density mask or with its own learned mask; paired gain and volumes won; 30~dB. Knee: real
k-space and scanner coil maps, 30k steps, 10 volumes. Brain: eight simulated coils, 10k steps; $120^3$ on 10
volumes, $60^3$ on all 150 validation volumes.}
\label{tab:acq}
\adjustbox{max width=\textwidth}{%
\begin{tabular}{llcccc}
\toprule
data & acquisition & $4\times$ & $8\times$ & $16\times$ & $32\times$ \\
\midrule
\multirow{3}{*}{SKM-TEA knee $128{\times}128{\times}80$} & protocol mask & 40.07 & 33.89 & 30.53 & 28.29 \\
 & learned mask & \textbf{42.45} & \textbf{35.52} & \textbf{31.41} & \textbf{28.95} \\
 & gain (won) & $+2.34$ (10/10) & $+1.57$ (10/10) & $+0.86$ (10/10) & $+0.67$ (10/10) \\
\midrule
\multirow{3}{*}{BraTS $120^3$} & protocol mask & 43.12 & 39.34 & 36.49 & 33.90 \\
 & learned mask & \textbf{45.82} & \textbf{41.04} & \textbf{37.06} & \textbf{34.05} \\
 & gain (won) & $+2.70$ (10/10) & $+1.69$ (10/10) & $+0.57$ (9/10) & $+0.16$ (7/10) \\
\midrule
\multirow{3}{*}{BraTS $60^3$ (150 volumes)} & protocol mask & 40.91 & 37.12 & \textbf{34.33} & \textbf{32.34} \\
 & learned mask & \textbf{45.54} & \textbf{39.24} & 33.93 & 31.68 \\
 & gain (won) & $+4.64$ (150/150) & $+2.12$ (150/150) & $-0.40$ (21/150) & $-0.66$ (6/150) \\
\bottomrule
\end{tabular}}
\end{table}

\paragraph{The reconstructor decides whether the acquisition pays.} LOUPE learns the pattern jointly with a
\emph{feed-forward} reconstructor, so the natural question is whether the gain above is just LOUPE in 3D.
Table~\ref{tab:acq2x2} trains all combinations of \{protocol, learned\} $\times$ \{feed-forward, warm CG\} at
$60^3$, everything else identical, on all 150 validation volumes. With a feed-forward reconstructor, learning
the acquisition \emph{hurts} at $4\times$ ($-1.74$~dB, won on $19/150$ volumes); with the projection the same
learned acquisition is worth $+4.64$~dB ($150/150$). A second seed of the feed-forward pair reads $-0.15/-1.80/-4.97/-2.57$~dB ($4/10$, then $0/10$;
10 volumes): with a feed-forward reconstructor the learned mask is at best neutral and unstable across seeds, while
its protocol-mask twin reproduces to $0.01$~dB. The feed-forward
model also fails to improve as the budget grows (it scores lower at $4\times$ than at $8\times$), which is the same fact from
the other side: a network that never enforces the measurement cannot cash in extra samples, so a
better sampling pattern buys it nothing. The two halves are not separable contributions. Two further controls
in the same table: the gain is not the hypernetwork's (with no conditioning at all it is $+4.55/+2.06/+0.02/-0.51$~dB, the same
curve), and it does not evaporate with a stronger reconstructor: with three
evaluations the learned mask wins at every rate, $+6.82/+5.22/+2.48/+0.94$~dB, on every one of the 150 volumes, and
with five it is still $+4.39/+4.53/+1.62$~dB at $4/8/16\times$ ($150/150$ each) and level at $32\times$. A second seed of the three-evaluation pair gives $+6.03/+4.64/+1.88/+0.24$~dB ($10/10$ volumes at
$4$ to $16\times$, $8/10$ at $32\times$; 10-volume subset). The mask head also interpolates: asked for accelerations it was
never trained on ($6/12/24\times$), it gains $+2.46/-0.03/-0.78$~dB with one evaluation and $+5.52/+3.20/+1.28$~dB
with three, on all 150 volumes ($150/150$ won wherever the gain is positive; Table~\ref{tab:unseenR}). The gain
peaks at three evaluations because the learned-mask model is by then close to the ceiling that $30$~dB measurement
noise sets ($48.9$ and $50.0$~dB at $4\times$ with three and five evaluations) while the protocol-mask model keeps
improving. A better
acquisition and a better reconstructor are complements, not substitutes. Training the sampler three times
longer, with a matched protocol-mask control also trained three times longer, moves both sides up by about
$0.5$~dB and leaves the gain unchanged ($+4.47/+2.17/-0.70/-1.36$~dB at 30k steps against $+4.48/+1.93/-0.64/-1.16$
at 10k; 10 volumes). Among untrained baselines on the same split (10
volumes), $\ell_1$-wavelet compressed sensing \citep{lustig2007sparse,beck2009fast} gives $36.21/30.23/27.49/26.05$ and CG-SENSE
$32.35/29.64/27.92/26.52$ ($33.26/30.60/28.94/27.59$ on all 150): the learned acquisition adds more on top of our
model at $4\times$ than sparsity adds on top of CG-SENSE.

\begin{table}[t]\centering\footnotesize\setlength{\tabcolsep}{4pt}
\caption{\textbf{Who can cash in a learned acquisition.} BraTS $60^3$, all 150 validation volumes, 30~dB, PSNR
(dB). All four reconstructors use the same backbone, data, masks and training; they differ in whether a
warm-started CG projection follows the network (rows 2 to 5) and in the number of evaluations. ``won'' counts the
volumes on which the learned mask beats the protocol mask.}
\label{tab:acq2x2}
\adjustbox{max width=\textwidth}{%
\begin{tabular}{llcccc}
\toprule
reconstructor & acquisition & $4\times$ & $8\times$ & $16\times$ & $32\times$ \\
\midrule
\multirow{3}{*}{feed-forward, 1 eval.\ (LOUPE form)} & protocol & 38.03 & 35.91 & 33.84 & 32.16 \\
 & learned & 36.30 & 36.64 & 32.88 & 29.20 \\
 & gain (won) & $-1.74$ (19) & $+0.73$ (134) & $-0.96$ (1) & $-2.95$ (0) \\
\midrule
\multirow{3}{*}{warm CG, 1 eval., no hypernetwork} & protocol & 40.85 & 37.07 & 34.30 & 32.31 \\
 & learned & 45.35 & 39.13 & 34.32 & 31.80 \\
 & gain (won) & $+4.55$ (150) & $+2.06$ (150) & $+0.02$ (76) & $-0.51$ (10) \\
\midrule
\multirow{3}{*}{warm CG, 1 eval., hypernetwork (ours)} & protocol & 40.91 & 37.12 & 34.33 & 32.34 \\
 & learned & 45.54 & 39.24 & 33.93 & 31.68 \\
 & gain (won) & $+4.64$ (150) & $+2.12$ (150) & $-0.40$ (21) & $-0.66$ (6) \\
\midrule
\multirow{3}{*}{warm CG, 3 eval., hypernetwork} & protocol & 42.13 & 37.45 & 34.19 & 31.90 \\
 & learned & \textbf{48.86} & \textbf{42.66} & \textbf{36.68} & \textbf{32.85} \\
 & gain (won) & $+6.82$ (150) & $+5.22$ (150) & $+2.48$ (150) & $+0.94$ (150) \\
\midrule
\multirow{3}{*}{warm CG, 5 eval., hypernetwork} & protocol & 45.68 & 39.78 & 35.79 & 33.22 \\
 & learned & \textbf{49.95} & \textbf{44.34} & \textbf{37.41} & \textbf{33.23} \\
 & gain (won) & $+4.39$ (150) & $+4.53$ (150) & $+1.62$ (150) & $+0.01$ (72) \\
\bottomrule
\end{tabular}}
\end{table}

\subsection{The cost-quality front against multi-step priors}\label{sec:front}
\begin{figure}[t]
\centering
\includegraphics[width=0.43\textwidth]{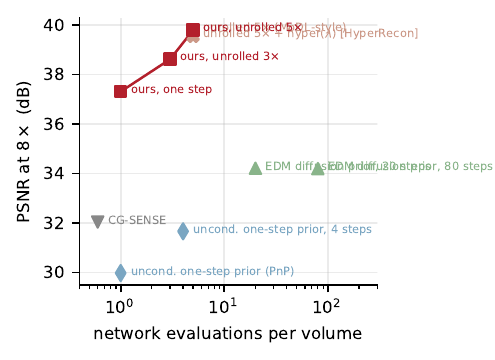}\hfill
\includegraphics[width=0.43\textwidth]{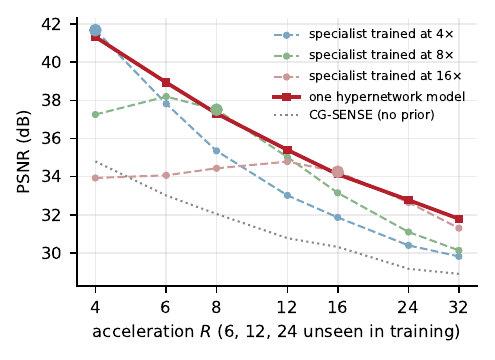}
\caption{\textbf{Left:} quality against network evaluations per volume ($8\times$, BraTS $60^3$). \textbf{Right:} one
hypernetwork model (red) against single-acceleration specialists (dashed; circles at their own rates); $6/12/24\times$
were seen by no model.}
\label{fig:main}
\end{figure}

Table~\ref{tab:front} places the reconstructor among prior-free, plug-and-play and supervised alternatives that
share its backbone, data, masks, noise realisations, solver and metrics; Fig.~\ref{fig:main} plots the front.
One evaluation beats the 20-step patch-diffusion prior at every rate, by a margin that grows with the
acceleration ($+1.5/+3.1/+3.1$~dB at $60^3$, $+1.4/+3.4/+4.5$ at $120^3$, $+0.8/+2.0/+2.9$ on knee), and the
80-step prior is within $0.2$~dB of the 20-step one, so the prior is saturated, not under-sampled; swept the other
way it loses $0.3/0.1/0.0$~dB at 10 steps and $0.9/0.3/0.0$~dB at 5 steps, still below our single evaluation at every
step count (App.~\ref{app:sweep}). On the acquisition split, evaluated on all 150 validation volumes, one evaluation is
$+2.03/+3.93/+4.29$~dB above the 20-step prior and three evaluations $+3.31/+4.28/+4.17$~dB, winning on $150/150$ volumes at every rate (Table~\ref{tab:front150});
on all 33 knee volumes the margins are $+0.82/+2.23/+3.07$ and $+1.94/+4.34/+4.89$~dB, $33/33$ at every rate
(Table~\ref{tab:knee33}). Three
readings. (i) \emph{The front, not the point.} One evaluation is the cheapest useful operating point: three and
five evaluations buy a further $+2.4/+1.3/+0.7$ and $+4.7/+2.8/+1.5$~dB, and the five-evaluation point beats the
20-step prior by $+6.2/+5.9/+4.6$~dB with four times fewer network calls. (ii) \emph{A negative result for
plug-and-play with a generative prior in the paired-data regime.} A supervised reconstructor with the same
backbone, trained through the same solver, is far ahead at lower cost; the generative prior keeps the advantage it was designed for, needing no paired data, but that advantage should be stated, not assumed. The corrected one-step
prior (App.~\ref{app:design}) is a capable plug-and-play denoiser, above CG-SENSE everywhere and above the
diffusion prior at $8$ to $16\times$ with four evaluations, so what the conditioned model buys is the $+6$~dB over
\emph{that}, not the rescue of a broken prior. The generative initialisation itself is worth little: from random
weights the same recipe gives $41.42/37.24/33.94$ against $41.33/37.31/34.12$ ($+0.19$~dB at $16\times$ only).
(iii) Amortising $\lambda$ alone, the HyperRecon form, is $0.6/0.25/0.15$~dB \emph{below} the same unrolled
model at a fixed $\lambda$, while the $(\lambda,R)$ hypernetwork is above its fixed-$\lambda$ counterpart at every
rate. Knee at $4\times$ is solver-bound (CG-SENSE already at $38.0$~dB; twenty warm CG iterations are worth
$+1.8$~dB there and nothing on brain). On $120^3$ the native model reaches $43.15/39.39/36.51$~dB and the $60^3$
model applied unchanged gives $41.67/36.95/33.93$, tying the natively trained prior at $4\times$ and beating it at
$8/16\times$, while implicit-kernel convolutions \citep{ma2023hyper} hurt in the same transfer
(App.~\ref{app:extended}).

\begin{table}[t]\centering\small
\caption{\textbf{The cost-quality front.} PSNR (dB), 10 volumes, 30~dB, best $\lambda$ per row; NFE = network
evaluations per volume (each diffusion step also runs a 40-iteration CG). All rows share backbone, data, masks,
noise, solver and metrics; ``unrolled'' = our recipe with $K$ iterations of [network $\to$ warm CG]. Brain: BraTS
$60^3$ stored-measurement split, CG-5; knee: SKM-TEA, CG-20.}
\label{tab:front}
\begin{tabular}{llcccc}
\toprule
 & method & NFE & $4\times$ & $8\times$ & $16\times$ \\
\midrule
\multirow{10}{*}{\rotatebox{90}{BraTS $60^3$}}
 & CG-SENSE (no prior)                                    & 0  & 34.80 & 32.04 & 30.31 \\
 & one-step MeanFlow prior, plug-and-play                  & 1  & 35.08 & 31.92 & 30.09 \\
 & one-step MeanFlow prior, plug-and-play                  & 4  & 38.80 & 34.36 & 31.42 \\
 & patch-diffusion prior, plug-and-play \citep{hu2024learning} & 20 & 39.83 & 34.22 & 31.04 \\
 & patch-diffusion prior, plug-and-play                    & 80 & 39.95 & 34.20 & 30.93 \\
 & \textbf{ours}, one evaluation, hyper$(\lambda,R)$       & 1  & 41.33 & 37.31 & 34.12 \\
 & \textbf{ours}, three evaluations, hyper$(\lambda,R)$    & 3  & 43.75 & 38.62 & 34.78 \\
 & unrolled $5\times$, hyper$(\lambda)$ (HyperRecon form)  & 5  & 44.72 & 39.54 & 35.41 \\
 & unrolled $5\times$, fixed $\lambda$ (MoDL/VarNet form)  & 5  & 45.33 & 39.79 & 35.56 \\
 & \textbf{ours}, five evaluations, hyper$(\lambda,R)$     & 5  & \textbf{46.01} & \textbf{40.13} & \textbf{35.60} \\
\midrule
\multirow{5}{*}{\rotatebox{90}{knee}}
 & CG-SENSE (no prior)                                    & 0  & 38.04 & 30.34 & 26.95 \\
 & one-step MeanFlow prior, plug-and-play                  & 4  & 38.50 & 31.62 & 27.46 \\
 & patch-diffusion prior, plug-and-play                    & 20 & 38.90 & 31.55 & 27.35 \\
 & \textbf{ours}, one evaluation, hyper$(\lambda,R)$       & 1  & 39.74 & 33.59 & 30.28 \\
 & \textbf{ours}, three evaluations, hyper$(\lambda,R)$    & 3  & \textbf{40.86} & \textbf{35.60} & \textbf{31.99} \\
\bottomrule
\end{tabular}
\end{table}

\IfFileExists{figs/fig_qual_knee_R8.pdf}{%
\begin{figure}[t]
\centering
\includegraphics[width=0.9\textwidth]{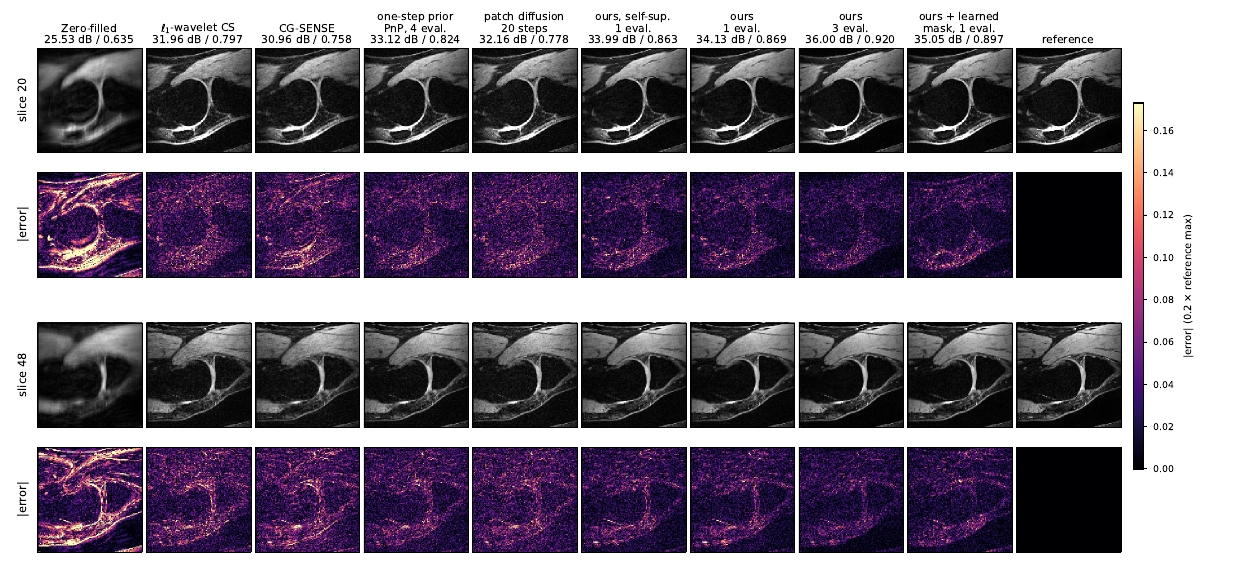}
\caption{\textbf{Reconstructions and error maps at $8\times$, SKM-TEA knee, volume 0, two slices.} Every column
is the same measurement (one mask, one noise draw, best $\lambda$ of the common grid) except the rightmost
method, which acquires with its own learned mask at the same budget; the PSNR/SSIM under each title is that of
the displayed volume. Error maps share one colour scale. At $4\times$ the knee is solver-bound and every method
is within $1$ to $4$~dB of the prior-free solver (App.~\ref{app:qual}); at $8\times$ the prior-based and the
measurement-conditioned reconstructions separate. The knee measurement is re-encoded from the reference with the
scanner coil maps, as in every knee table.}
\label{fig:qual_main}
\end{figure}
}{%
\begin{figure}[t]
\centering
\includegraphics[width=\textwidth]{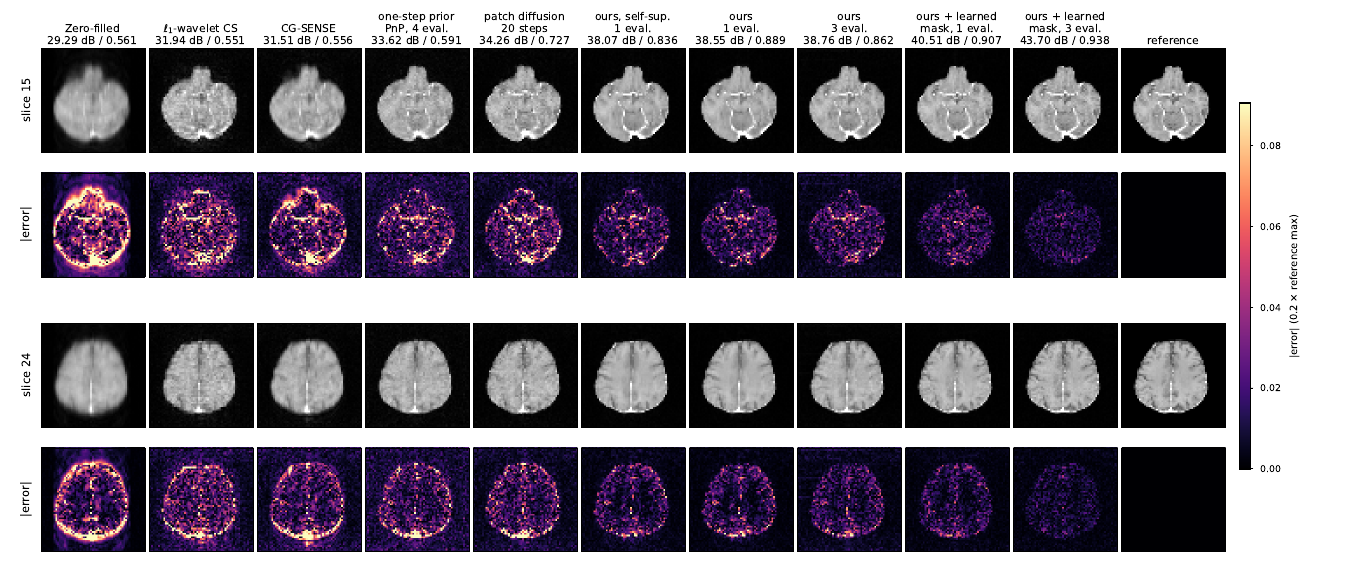}
\caption{\textbf{Reconstructions and error maps at $8\times$, BraTS $60^3$, volume 3, two axial slices.} Every column
is the same measurement (one mask, one noise draw, best $\lambda$ of the common grid) except the two rightmost
methods, which acquire with their own learned mask at the same budget; the PSNR/SSIM under each title is that of
the displayed volume. Error maps share one colour scale. Further accelerations, volumes and the knee set are in
App.~\ref{app:qual}.}
\label{fig:qual_main}
\end{figure}
}
\subsection{What is worth amortising}\label{sec:amort}
Table~\ref{tab:spec} (App.~\ref{app:extended}) evaluates one hypernetwork model against specialists of the same recipe trained at a
single acceleration, at the trained rates and at three rates no model saw. One model is within
$0.36/0.20/0.10$~dB of the specialists at their own rates and $2$ to $7$~dB above them away from those rates (a
$16\times$ specialist evaluated at $4\times$ loses $7.4$~dB). Against a specialist trained over the same $R$ range
but blind to it, the hypernetwork is $+0.28/+0.19/+0.15$~dB better ($7/10$, $10/10$, $10/10$ volumes; at 30k steps
$+0.30/+0.19/+0.16$ with intervals excluding zero), and at five evaluations $+0.33$~dB at $8\times$. The value
comes from $R$: specialists trained at $\lambda=0.003$, $0.03$ and $0.3$ are indistinguishable ($\le0.03$~dB),
so with a warm-started solver the training-time data-consistency weight is immaterial and amortising it is trivial, consistent with the HyperRecon-form row of Table~\ref{tab:front}. The one weak spot, $6\times$ between two training rates, is an artefact of a Fourier-feature input encoding: a
plain MLP (the HyperMorph form) gives $38.94$~dB there, above both neighbouring specialists, loses nothing at the
trained rates, and continuous $R\sim\log\mathcal U[3,40]$ closes the hole equally (App.~\ref{app:extended}); we use
the plain MLP. A user-facing dial we hoped to expose, a fidelity/sharpness loss trade-off $\alpha$ given to the hypernetwork,
moves the reconstruction by $0.02$~dB across its range (App.~\ref{app:extended}): a hypernetwork
is only as interesting as the hyperparameter it is given, and a loss reweighting does not survive the
data-consistency step.

\subsection{Without fully sampled targets}\label{sec:ssl}
On the real knee data, the reconstructor trained with no target at all (Sec.~\ref{sec:ssl-method}) gives
$39.93/33.41/29.99/27.67$~dB against $39.92/33.64/30.29/28.06$ for its supervised twin: indistinguishable at
$4\times$ ($5/10$ volumes, interval $[-0.09,+0.03]$) and at most $0.4$~dB behind at $32\times$, where the held-out
half of a very sparse acquisition is itself a poor teacher; both are far above CG-SENSE ($38.0/30.3/27.0$) and the diffusion prior ($38.9/31.6/27.4$); on all 33 volumes the
difference is $-0.02/-0.20/-0.29/-0.40$~dB. The fully sampled archive is therefore not required. Combining this
with the learned acquisition (no fully sampled data \emph{and} a pattern chosen by the model) fails, badly
and instructively: at $60^3$ the self-supervised learned mask gives $33.62/31.96/30.89/30.58$ against
$40.42/36.78/34.14/32.23$ for self-supervision with the protocol mask (150 volumes, $0/150$ won at every rate). The masks say what happened: the fraction
of sampled locations with sampled neighbours is $0.94/0.90/0.70/0.55$ for the self-supervised sampler against
$0.20/0.20/0.27/0.46$ for the supervised one. It acquires in clumps, which is the optimum of the objective it was given (a held-out sample surrounded by acquired ones is nearly free to predict) and close
to the worst thing
for reconstruction. When the model both sets the exam and sits it, it optimises the exam; a self-supervised loss
to be combined with a learned acquisition needs a held-out set the sampler cannot influence.

\subsection{Negative results, with their mechanisms}\label{sec:neg}
\textbf{Subject-adaptive acquisition.} Acquiring a fixed low-frequency block first and letting a convolutional
policy choose the remaining lines from the energy it measured is far \emph{worse} than the static learned mask
($-10.2/-6.6/-1.9/-0.7$~dB at $60^3$, $-10.0/-7.3/-3.7/-2.5$ at $120^3$, $-10.5/-5.4/-2.6/-1.9$ on real knee,
losing every volume): conditioning on measured energy makes the policy greedy for the high-energy central lines it
already has. Energy is not information. \textbf{Stochastic one-step sampling} collapses: with the clean condition
visible and an MSE loss the network learns to ignore the injected noise, the coverage of a nominal-90\% interval
falls from $0.41$ to $0.03$ to $0.12$ as training proceeds, and the spread was under-training, not posterior
uncertainty (App.~\ref{app:design}). \textbf{One model across anatomies} costs $0.8$~dB on brain and nothing on
knee; \textbf{implicit-kernel convolutions} do not transfer across resolution, the patch/thumbnail parametrisation
does (App.~\ref{app:extended}).

\section{Discussion and conclusion}
\textbf{The acquisition is the operating point worth learning}, and only a reconstructor that enforces the
measurement can cash it in ($-1.7$~dB with a feed-forward network, $+4.6$~dB with a five-iteration projection).
\textbf{Where the network sits relative to the solver decides everything}: every failed design failed because the
network's estimate could not reach the loss (a cold truncated CG), the measurement could not reach the network (a
noised input), the network's output was scaled to zero (the MeanFlow jump at $t\to0$), or the sampler could reach
the loss it was scored by (self-supervised learned acquisition). \textbf{Amortise what changes the answer}: the
acceleration and the acquisition, not the data-consistency weight or a loss trade-off. A MeanFlow-parametrised,
measurement-conditioned 3D reconstructor with an amortised operating point beats the 20-step patch-diffusion prior
it was initialised from at every acceleration on brain and knee, one network covers $4$ to $32\times$, and the scan
can supervise itself. What remains open is uncertainty (a one-step model trained with a squared loss has none to give) and a
self-supervised objective that can be combined with a learned acquisition.

\subsubsection*{Reproducibility statement}
Every dataset is built from a public source by a script in the repository, with masks and noise drawn from
generators seeded per volume; every number in the paper is backed by a per-volume result table (App.~\ref{app:log})
and the code, configurations and checkpoints are released. Rebuilding both datasets from the public archives with the released scripts and re-evaluating the trained models
reproduces the published brain numbers to $0.0000$~dB and the knee numbers to within $0.005$~dB across four
accelerations (App.~\ref{app:protocol}).

\subsubsection*{Ethics statement}
Both datasets are public and de-identified. The brain results use simulated coil sensitivities and are marked
as such; the knee results use real acquisitions. Learned sampling patterns change what a scanner measures, and
nothing here has been validated prospectively or for clinical use.

\subsubsection*{Acknowledgements}
All experiments were run on NVIDIA Titan RTX GPUs. We thank the SKM-TEA and BraTS teams for releasing the data.

\bibliographystyle{iclr2027_conference}
\bibliography{refs}

\appendix
\section{Protocol details and the reproducibility test}\label{app:protocol}
\paragraph{Protocol.} PSNR on the magnitude over the whole volume with the target maximum as range,
Wang-SSIM per axial slice, and a $\lambda$ sweep for every method. Data: SKM-TEA knee
\citep{desai2022skm} at quarter resolution ($128{\times}128{\times}80$) from the fully sampled raw
acquisitions, and BraTS \citep{baid2021rsna,de20242024} at $60^3$ and $120^3$ with eight simulated coils. Undersampling
uses a variable-density Gaussian mask on the phase-encode plane with a fully sampled centre; complex
Gaussian k-space noise is added at a nominal 30~dB. \emph{How that noise is normalised matters, and the
two natural choices differ by more than most method gaps.} Drawing the noise over the full k-space grid,
scaling it to the nominal SNR against $\|y\|$ on the full grid, and masking afterwards leaves only a
fraction $1/R$ of the noise energy on the sampled points, so the effective SNR on the data is
$30+10\log_{10}R$~dB ($36/39/42$~dB at $4/8/16\times$); measuring the noise power on the sampled points
gives a true 30~dB. We use the sampled-point convention throughout, report the full-grid convention
where marked (``light noise'') for readers reproducing either, and apply one convention to every arm of
a comparison. Practitioners should state which they use: it is worth $1$ to $2$~dB.

\paragraph{Reproducibility, tested.} Every dataset here is built from a public source by a script in the
repository, with the masks and the noise drawn from generators seeded per volume. Rebuilding both datasets from the
public archives and re-evaluating the trained models reproduces every published brain number to $0.0000$~dB across
four accelerations and ten volumes, and the knee numbers to $+0.004/-0.001/+0.002/+0.002$~dB (protocol mask) and
within $0.001$~dB (learned mask) at $4/8/16/32\times$, the residual being floating-point arithmetic. All models were
trained on single NVIDIA Titan RTX GPUs. Checkpoints and the per-volume result tables behind every number in this
paper are released with the code.

\paragraph{What the measurement is, exactly.} Both datasets give us an image and a set of coil sensitivities,
and we form the measurement ourselves: $y = M\,\mathcal{F}Sx^\star + \varepsilon$. For SKM-TEA the image and the
sensitivities come from the scanner (fully sampled acquisitions, vendor coil maps); because we work at quarter
resolution we re-encode with the downsampled maps, so the noiseless part of $y$ is consistent with $x^\star$ by
construction. This is worth stating because it is the standard practice in this literature and it silently
inflates every iterative reconstructor. We measured it: with the same volumes, masks and solver, a
\emph{prior-free} CG-SENSE scores $34.78/30.77/28.55/26.96$~dB at $4/8/16/32\times$ on noiselessly consistent
data and $32.35/29.64/27.92/26.52$ once $30$~dB measurement noise is present, an inflation of
$+2.43/+1.13/+0.63/+0.44$~dB that costs nothing to obtain and is larger, at $4\times$, than many published
method gaps. Adding noise at a stated SNR is what makes the problem an estimation problem rather than a
noiseless inversion, and we do it everywhere. For BraTS the sensitivities are simulated (eight analytic maps), which we mark as such
wherever those numbers appear.

\section{The one-step prior and the design path}\label{app:design}
\subsection{Training the MeanFlow patch prior}\label{app:prior}
Rectified flow: $z_t = (1-t)x + t\epsilon$, $v = \epsilon - x$. We model the average velocity
$u(z_t, r, t) = \frac{1}{t-r}\int_r^t v\,d\tau$, so $z_r = z_t - (t-r)\,u(z_t,r,t)$ is exact, trained with the
identity $u = v - (t-r)\,\frac{d}{dt}u$ (stop-gradient right-hand side). The network sees a $P^3$ patch of $z_t$,
the whole $z_t$ pooled to $P^3$, and the patch position; the thumbnail's tangent $\mathrm{pool}(v)$ enters the
Jacobian-vector product. Training follows \citet{geng2026mean}: sample $(r,t)$ with $t\sim$ logit-normal and $r=t$
with probability $p_{\mathrm{equal}}$ (then $u$ reduces to the instantaneous velocity and the loss to rectified
flow), regress $u_\theta$ onto the stop-gradient target $v - (t-r)\,\frac{d}{dt}u_\theta$ where the total
derivative is one forward-mode JVP along $(v, \mathrm{pool}(v), 0, 1)$ (the thumbnail moves with the patch) and weight each sample adaptively by $1/(\|\text{err}\|^2+c)^{p}$ with the weight detached. We found
$p_{\mathrm{equal}}=0.5$ better than the paper's $0.25$ by $0.3$ to $1.1$~dB in downstream reconstruction, and run
the flow on $x/\sigma_{\mathrm{data}}$ (MRI intensities normalised to $[0,1]$ have $\sigma\approx0.1$; without
rescaling the path is noise-dominated at every $t$). Used on its own as a plug-and-play denoiser the prior
starts at $(1-t_s)\,A^{\!H}y + t_s\epsilon$, jumps once to $\hat x$, and runs one CG data-consistency solve.
The corrected training settings referred to in the main text are: adaptive constant $c=0.01$, logit-normal mean
$-0.4$, Adam $\beta_2=0.95$, no mixed precision.

\subsection{What failed on the way, and why}
The following record, in the order the experiments were run, is what fixed the design of Sec.~\ref{sec:method}.

\begin{itemize}
  \item \textbf{Multi-step saturates early.} On BraTS-24 (simulated 8-coil, 30~dB k-space noise,
  10 volumes) the EDM patch prior at 20 steps is within 0.2~dB of 80 steps at every rate
  ($120^3$: 41.73/36.03/32.03 vs 41.95/36.15/32.07~dB at $4/8/16\times$), at $1/4$ of the cost.
  \item \textbf{The prior is worth a lot here.} EDM prior vs.\ the prior-free CG-SENSE control on the
  identical code path: $+6.7/+4.6/+2.3$~dB ($120^3$) and $+5.9/+2.7/+1.2$~dB ($60^3$) at $4/8/16\times$: unlike knee at $4\times$, simulated-coil brain reconstruction is prior-bound already at $4\times$.
  \item \textbf{An unconditional one-step flow prior does not de-alias.} Used as a plug-and-play
  denoiser from a warm start, the MeanFlow one-step estimate has ZF-level PSNR ($26.7$ vs ZF $27.2$~dB,
  $60^3$, $4\times$): it removes the injected noise but has never seen aliasing. On the BraTS $60^3$
  evaluation set (30~dB, 10 vols, $\lambda$ and $t_{\mathrm{start}}$ tuned) it lands
  \emph{below} the prior-free CG-SENSE control at NFE$=1$ ($32.96/30.96/29.59$ vs $34.80/32.04/30.31$~dB
  at $4/8/16\times$) and only $+0.6/+0.5/+0.2$~dB above it at NFE$=4$, while the multi-step EDM prior on
  the same set is $\approx+6$~dB above the control. The best warm start is the smallest one tried
  ($t_{\mathrm{start}}=0.1$; monotone), i.e.\ the less the one-step prior moves the zero-filled image,
  the better. Multi-step PnP works because each step re-noises heavily. \emph{One step must be measurement-conditioned}, trained from the zero-filled image through the data-consistency step
  (Sec.~\ref{sec:hyper}); this is the conditioned reconstructor of Sec.~\ref{sec:method}.
  \item \textbf{A one-step prior is a fine denoiser once it is trained correctly; our first negative was a tuning artefact.} With the settings we started from, the unconditional one-step prior used plug-and-play sits
  below CG-SENSE ($32.96$ vs.\ $34.80$~dB at $4\times$, one evaluation). Retraining it with the corrected
  MeanFlow settings (adaptive constant $0.01$, logit-normal mean $-0.4$, Adam $\beta_2=0.95$, no mixed
  precision) gives $35.08/31.92/30.09$ at one evaluation and $38.80/34.36/31.42$ at four, above CG-SENSE everywhere, and at $8$ and $16\times$ \emph{above the 20-step diffusion prior} at a fifth of its cost. We
  report the correction rather than the first number: what the measurement-conditioned model buys is not
  rescuing a broken prior, it is the $+6$~dB over this corrected baseline.
  \item \textbf{The generative initialisation is worth little, and only where data is scarcest.} Training the
  identical conditioned reconstructor from random weights instead of from the one-step prior gives
  $41.42/37.24/33.94$~dB against $41.33/37.31/34.12$ at $4/8/16\times$ (10 volumes, same recipe, same steps):
  $-0.08$~dB at $4\times$ (2/10 volumes, confidence interval including zero), $+0.08$ at $8\times$, and $+0.19$
  at $16\times$ (10/10 volumes, interval excluding zero). A better prior does not change the picture: initialising from a one-step prior trained with the
  corrected MeanFlow settings (adaptive constant $0.01$, logit-normal mean $-0.4$, Adam $\beta_2=0.95$, no mixed precision), which is itself $2$~dB stronger as a plug-and-play
  denoiser, gives $41.41/37.41/34.29$~dB, that is $+0.00/+0.19/+0.36$~dB over random weights ($3/10$, $8/10$ and
  $10/10$ volumes). The prior helps exactly where the measurement determines least, and by a few tenths of a
  decibel. We therefore do not claim the generative model as the source of the method's accuracy; what carries
  it is the measurement conditioning, the training through the solver, and the amortised operating point.
  \item \textbf{End-to-end through a truncated solver needs a warm start (negative).} Fine-tuning
  the one-step model through a \emph{cold-started} 5-iteration CG (the PnP proximal step, $z_0=0$)
  leaves the training loss flat for 10k steps ($0.0006$ to $0.0007$): the truncated solver swallows the
  gradient. The resulting models are indistinguishable from their initialisation: the hypernetwork
  model and a fixed-$\lambda$ model agree within $0.1$~dB at every $\lambda$ and the $\lambda$-curve
  they trace is the 40-iteration CG's, not the network's. On the full 10-volume $60^3$ set (30~dB,
  best $\lambda$ per rate) the v1 hypernetwork model lands \emph{on} the prior-free CG-SENSE control:
  $34.7/31.8/30.2$ vs $34.8/32.0/30.3$~dB at $4/8/16\times$ (fixed-$\lambda$ v1 models: same); the network's estimate contributes nothing once the cold CG has run. The second design starts the CG at $\hat x$ and supervises $\hat x$ directly; the third fixes the parametrisation (next item).
  \item \textbf{A one-step \emph{prior} is the wrong parametrisation for a one-step \emph{reconstructor}
  (measured, drives the final design).} Reading out $\hat x$ itself (before any CG) on two $60^3$ volumes at
  $4\times$: the MeanFlow initialisation gives $26.5$~dB ($\approx$ ZF, $27.2$), v1 after 10k steps $27.5$,
  v2 after 500 steps $27.7$, while the same estimates after a 5-iteration warm CG are $32.6$ / $34.3$ /
  $34.4$~dB. Two causes. (i) At $t_s=0.5$ the input $0.5\,x_{\mathrm{zf}}+0.5\,\epsilon$ has noise larger
  than signal: the measurement is destroyed at the input and no amount of training recovers it (the
  stochastic dial of P3-style samplers costs fidelity for exactly this reason). (ii) The MeanFlow jump
  $\hat x = z - t_s u$ scales the network's contribution by $t_s$, so the deterministic limit $t_s=0$ is the
  identity on the zero-filled image; our earlier deterministic controls were literally ZF~+~CG. The
  conditioned reconstructor of Sec.~\ref{sec:hyper} (clean ZF as condition channels, jump floor) removes both.
  \item \textbf{The conditioned one-step reconstructor learns (preliminary, in-training validation).} With the
  clean zero-filled image as a condition and the jump floor, the training loss drops $4\times$ in 500 steps
  and the one-volume validation (native $4\times$, no added noise, 5-iteration warm CG) reaches $40.3$~dB for
  the hypernetwork model and $41.1$~dB for a fixed-$\lambda$ model after 1000 steps, against $36.1$~dB for
  the noise-limited v2 design (flat from 1000 to 3000 steps) and CG-SENSE-level for v1; at 5000 steps
  $43.2$ / $43.0$~dB and still rising. The stochastic model with the MeanFlow-form jump ($\hat x = z - s u$)
  plateaued at $36.3$~dB when started at $t_s=0.5$, the v2 level: at large $t_s$ it has to reproduce the
  injected noise exactly to cancel it. Taking the residual from the clean condition instead (Sec.~\ref{sec:hyper})
  removes that requirement but recovers only $0.25$~dB at matched steps: the binding constraint is that the
  network must learn to \emph{ignore} the noisy channel and route through the clean condition, and any leakage
  injects noise into the output. An MSE-trained stochastic one-step model therefore converges toward the
  deterministic one; we report its spread-vs-fidelity curve as a limitation rather than as posterior sampling.
  \item \textbf{Headline (10 volumes, BraTS $60^3$ set, 30~dB): the conditioned one-step reconstructor beats
  the 20-step diffusion prior at every rate, and one hypernetwork model beats its own specialist.} Deterministic
  one-step, five warm CG iterations: hypernetwork $(\lambda,R)$ $41.33/37.31/34.16$~dB vs EDM-20 $39.83/34.22/31.04$
  and CG-SENSE $34.80/32.04/30.31$ at $4/8/16\times$ (table below). The advantage grows with $R$ ($+1.5\to+3.1$~dB),
  i.e.\ exactly where the unconditional prior + PnP is weakest. The $\lambda$ dependence of the hypernetwork model is
  flat ($\le0.02$~dB over $\lambda\in[10^{-3},0.03]$), which is what makes test-time selection  safe. With the
  clean condition visible the stochastic variant costs $\le0.4$~dB from $t_s=0$ to $0.5$ (v3, $36.5\to36.1$ at
  $4\times$) instead of $1.8$~dB without it (v2), but sits $\sim5$~dB below the deterministic model at 10k steps.
  The same hypernetwork model also covers $32\times$ (trained range $4$ to $32\times$): $31.83$~dB vs $31.76$ for the
  $\lambda{=}0.03$ specialist and $28.90$ for CG-SENSE.
  \item \textbf{Stochastic one-step UQ: the dial collapses with training (negative).} Eight one-step samples
  of the conditioned stochastic model (warm CG, $\lambda{=}0.01$, 10 volumes). After 10k steps the empirical
  coverage of the nominal-90\% interval $|\text{err}| \le 1.645\,\sigma$ still grows with the dial
  ($0.06/0.22/0.26/0.41$ at $t_s=0.05/0.2/0.5/0.7$, $4\times$) at a cost of $0.5$~dB; after 30k steps the same
  model reaches the deterministic model's fidelity at every $t_s$ ($41.35\to41.18$~dB from $t_s=0$ to $0.5$ at
  $4\times$) and the coverage falls to $0.03$ to $0.12$: with the clean condition visible and an MSE loss the
  network learns to ignore the injected noise, and the spread that remained earlier was under-training, not
  posterior uncertainty. Stochastic one-step sampling therefore gives no usable uncertainty in this design; we
  report it as a negative result.
  \item \textbf{$120^3$ confirms the $60^3$ result with a larger margin.} Natively trained at $120^3$ the hypernetwork
  one-step model reaches $43.15/39.39/36.51$~dB (SSIM $0.902/0.863/0.853$) against $41.73/36.03/32.03$ for the 20-step
  EDM prior and $35.29/31.53/29.74$ for CG-SENSE (10 volumes, 30~dB): $+1.4/+3.4/+4.5$~dB.
  \item \textbf{Zero-shot resolution transfer comes for free with the patch parametrisation.} The $60^3$-trained
  hypernetwork model evaluated unchanged on $120^3$ volumes gives $41.67/36.95/33.93$~dB: it ties the natively-trained
  20-step EDM prior at $4\times$, beats it at $8/16\times$, and sits $1.5$ to $2.6$~dB below native training.
  Implicit-kernel convolutions with tap re-sampling (hyper-convolutions) hurt in the same transfer, on both the
  unconditional prior and the conditioned reconstructor ($-1.5$~dB natively, $-2.2$~dB zero-shot at 3 taps,
  $-12$~dB at 5 taps): a negative result for kernel-level resolution adaptation in this setting.
\end{itemize}

\section{Extended results}\label{app:extended}
\subsection{One hypernetwork against single-rate specialists}
\begin{table}[h]\centering\footnotesize\setlength{\tabcolsep}{4pt}
\caption{\textbf{One hypernetwork against single-rate specialists} (BraTS $60^3$, PSNR dB, 10 volumes, 30~dB).
$6/12/24\times$ were seen by no model.}
\label{tab:spec}
\begin{tabular}{lccccccc}
\toprule
 & $4\times$ & $6\times$ & $8\times$ & $12\times$ & $16\times$ & $24\times$ & $32\times$ \\
\midrule
specialist $4\times$  & \textbf{41.69} & 37.82 & 35.35 & 33.02 & 31.86 & 30.39 & 29.82 \\
specialist $8\times$  & 37.26 & 38.20 & \textbf{37.51} & 35.03 & 33.15 & 31.09 & 30.13 \\
specialist $16\times$ & 33.93 & 34.07 & 34.44 & 34.79 & \textbf{34.26} & 32.64 & 31.30 \\
one hypernetwork $(\lambda,R)$, Fourier features & 41.33 & 37.04 & 37.31 & 35.40 & 34.16 & 32.73 & \textbf{31.83} \\
one hypernetwork $(\lambda,R)$, plain MLP & 41.33 & \textbf{38.94} & 37.31 & \textbf{35.40} & 34.12 & \textbf{32.77} & 31.79 \\
CG-SENSE & 34.80 & 33.01 & 32.04 & 30.77 & 30.31 & 29.16 & 28.90 \\
\bottomrule
\end{tabular}
\end{table}
\subsection{The protocol-matched brain front in full}
\paragraph{Main result on the BraTS $60^3$ evaluation set (10 volumes, 8-coil, k-space SNR 30~dB).}
All rows share the forward operator, masks, noise realisations, CG code and metrics; $\lambda$ is the best of the
same grid for every row. NFE = network evaluations per volume (each EDM step also runs a 40-iteration CG).
\begin{center}\footnotesize
\adjustbox{max width=\textwidth}{%
\begin{tabular}{lcccc}
\toprule
 & NFE & $4\times$ & $8\times$ & $16\times$ \\
\midrule
CG-SENSE (prior-free)                       & 0  & 34.80 / 0.678 & 32.04 / 0.631 & 30.31 / 0.611 \\
Unconditional one-step flow prior + CG      & 1  & 32.96 / 0.699 & 30.96 / 0.656 & 29.59 / 0.627 \\
EDM patch prior + CG (20 steps)             & 20 & 39.83 / 0.845 & 34.22 / 0.762 & 31.04 / 0.727 \\
EDM patch prior + CG (80 steps)             & 80 & 39.95 / 0.847 & 34.20 / 0.761 & 30.93 / 0.724 \\
\midrule
Ours, one-step conditioned, fixed $\lambda{=}0.03$ (CG-5) & 1 & 41.06 / 0.867 & 37.12 / 0.852 & 34.01 / 0.837 \\
Ours, one-step conditioned, \textbf{hypernetwork} $(\lambda, R)$ (CG-5) & 1 & \textbf{41.33 / 0.886} & \textbf{37.31 / 0.858} & \textbf{34.16 / 0.841} \\
Ours, one-step conditioned, \textbf{hypernetwork} $(\lambda, R)$ (CG-20) & 1 & 41.29 / 0.876 & 37.27 / 0.844 & 34.14 / 0.828 \\
\bottomrule
\end{tabular}}
\end{center}
PSNR~/~Wang-SSIM. Both of our models: $60^3$, 10k end-to-end steps from the same MeanFlow initialisation, one
network pass plus a warm-started CG projection at test time (5 iterations = the training setting, used in all
ablation tables; 20 iterations = the setting we report against other methods, see the knee results). A second
training seed of the hypernetwork model gives $41.36/37.32/34.16$~dB (10 volumes; on all 200 volumes three seeds (two with the Fourier encoding, one plain-MLP) give
$41.16/37.17/34.07$, $41.20/37.18/34.07$ and $41.20/37.18/34.03$~dB). On all 200 volumes of the set (CG-5, mean $\pm$ std over
volumes): CG-SENSE $34.54{\pm}1.39$ / $31.82{\pm}1.36$ / $30.08{\pm}1.38$, EDM-20
$39.64{\pm}1.35$ / $34.06{\pm}1.25$ / $30.87{\pm}1.20$, ours $41.16{\pm}1.59$ / $37.17{\pm}1.58$ /
$34.07{\pm}1.51$~dB (SSIM $0.881/0.854/0.836$ vs.\ $0.835/0.752/0.717$ for EDM-20): $+1.5/+3.1/+3.2$~dB. The single hypernetwork model is evaluated at
every $\lambda$ of the grid and is flat to $0.02$~dB over $\lambda\in[10^{-3},3\cdot10^{-2}]$; it is $0.15$ to $0.27$~dB
\emph{above} the specialist trained at $\lambda=0.03$ and $+1.5/+3.1/+3.1$~dB above the 20-step diffusion prior
at $1/20$ of the network cost. Under the light-noise convention (effective $36/39/42$~dB, the numbers
that belong in their tables) the same three arms give, on all 200 volumes, CG-SENSE $35.33{\pm}1.37$ /
$32.36{\pm}1.34$ / $30.47{\pm}1.36$, EDM-20 $40.72{\pm}1.35$ / $34.73{\pm}1.31$ / $31.39{\pm}1.20$ and ours (CG-20)
$41.76{\pm}1.60$ / $37.53{\pm}1.59$ / $34.27{\pm}1.52$~dB at $4/8/16\times$ (SSIM $0.889/0.872/0.843$ vs.\
$0.870/0.731/0.742$): the lighter noise helps the multi-step prior most at $4\times$, and the one-step model stays
$+1.0/+2.8/+2.9$~dB ahead. At $120^3$
under the same convention: CG-SENSE $36.28/32.27/30.25$, EDM-20 $43.36/37.45/32.88$, ours (CG-20)
$44.11/40.07/36.97$ ($+0.8/+2.6/+4.1$~dB). Two further readings of the same experiment. (i) Specialists trained at
$\lambda=0.003$, $0.03$ and $0.3$ are indistinguishable ($\le0.03$~dB at every evaluation $\lambda$; even the
$\lambda{=}0.3$ specialist evaluated at $0.3$ gives $40.14$~dB against the hypernetwork's $40.56$): with a
warm-started five-iteration CG the training-time data-consistency weight is immaterial, so amortising over
$\lambda$ is trivial and the hypernetwork's $+0.3$~dB over every specialist comes from the other input it
receives, the acceleration $R$ (the specialists are trained over the same $R$ range but are blind to it). Per volume, the
hypernetwork model beats the specialist in $7/10$, $10/10$ and $10/10$ volumes at $4/8/16\times$ (paired
mean $+0.28/+0.19/+0.15$~dB). Training both three times longer (30k steps) moves them together
(hypernetwork $41.63/37.43/34.18$, specialist $41.34/37.24/34.02$~dB) and makes the margin significant at every
rate: $+0.30/+0.19/+0.16$~dB, $10/10$ volumes each, 95\% CIs $[+0.05,+0.54]$, $[+0.03,+0.35]$, $[+0.02,+0.29]$.
On knee the same holds with a smaller margin (hypernetwork $37.94/33.10/30.16$ vs.\ its $\lambda{=}0.03$
specialist $37.91/32.98/30.08$, CG-5). The amortisation that matters is therefore over $R$. Single-rate specialists (same recipe, trained at one
$R$) against the one hypernetwork model, PSNR at each evaluation rate ($6/12/24\times$ were seen by none of
the models; true 30~dB, 10 volumes):
\begin{center}\footnotesize
\adjustbox{max width=\textwidth}{%
\begin{tabular}{lccccccc}
\toprule
 & $4\times$ & $6\times$ & $8\times$ & $12\times$ & $16\times$ & $24\times$ & $32\times$ \\
\midrule
specialist $4\times$  & \textbf{41.69} & 37.82 & 35.35 & 33.02 & 31.86 & 30.39 & 29.82 \\
specialist $8\times$  & 37.26 & \textbf{38.20} & \textbf{37.51} & 35.03 & 33.15 & 31.09 & 30.13 \\
specialist $16\times$ & 33.93 & 34.07 & 34.44 & 34.79 & \textbf{34.26} & 32.64 & 31.30 \\
one hypernetwork $(\lambda,R)$ & 41.33 & 37.04 & 37.31 & \textbf{35.40} & 34.16 & \textbf{32.73} & \textbf{31.83} \\
CG-SENSE & 34.80 & 33.01 & 32.04 & 30.77 & 30.31 & 29.16 & 28.90 \\
\bottomrule
\end{tabular}}
\end{center}
One model is within $0.36/0.20/0.10$~dB of the three specialists at their own rates and $2$ to $7$~dB above
them away from those rates (a $16\times$ specialist evaluated at $4\times$ loses $7.4$~dB to it). Its one weak
spot is $6\times$, between two training rates, where both neighbouring specialists beat it by up to $1.2$~dB:
the hypernetwork trained on the discrete set $\{4,8,16,32\}$ interpolates poorly there. The ablation below
locates the cause in the hypernetwork's \emph{input encoding}: our default lifted $(\log\lambda,\log_2R)$ with 16
Fourier features (frequencies up to $2^{15}\pi$), which makes the embedding essentially unrelated between
neighbouring training rates. A plain MLP on the normalised inputs (the HyperMorph form) gives $38.94$~dB at
$6\times$, above both neighbouring specialists, while losing nothing at the trained rates; 4 features sit in
between. The same table separates the two inputs: a hypernetwork over $R$ alone reproduces the $(\lambda,R)$
model at every trained rate (the $\lambda$ input carries nothing, consistent with the specialists above), and a
hypernetwork over $\lambda$ alone, i.e.\ a model blind to $R$, loses $0.3/0.2/0.15$~dB at $4/8/16\times$: the
hypernetwork's gain is the $R$ conditioning.
\begin{center}\footnotesize
\adjustbox{max width=\textwidth}{%
\begin{tabular}{lccccccc}
\toprule
hypernetwork input / encoding & $4\times$ & $6\times$ & $8\times$ & $12\times$ & $16\times$ & $24\times$ & $32\times$ \\
\midrule
$(\lambda,R)$, 16 Fourier features (default above) & 41.33 & 37.04 & 37.31 & 35.40 & 34.16 & 32.73 & 31.83 \\
$(\lambda,R)$, 4 Fourier features & 41.37 & 38.24 & 37.32 & 35.48 & 34.14 & 32.86 & 31.81 \\
$(\lambda,R)$, plain MLP & 41.33 & \textbf{38.94} & 37.31 & 35.40 & 34.12 & 32.77 & 31.79 \\
$R$ only, 16 Fourier features & 41.32 & 35.94 & 37.29 & 35.41 & 34.11 & 32.83 & 31.80 \\
$\lambda$ only (blind to $R$), 16 Fourier features & 40.99 & 38.69 & 37.10 & 35.25 & 34.01 & 32.76 & 31.78 \\
$(\lambda,R)$, 16 Fourier features, continuous $R\sim\log\mathcal U[3,40]$ & 41.27 & 38.95 & 37.31 & 35.42 & 34.13 & 32.87 & 31.80 \\
$(\lambda,R)$, 4 Fourier features, continuous $R$ & 41.33 & 38.95 & 37.31 & 35.46 & 34.15 & 32.87 & 31.80 \\
$(\lambda,R)$, plain MLP, continuous $R$ & 41.32 & 38.96 & 37.31 & 35.45 & 34.15 & 32.85 & 31.81 \\
\bottomrule
\end{tabular}}
\end{center}
Training with a continuous $R\sim\log\mathcal U[3,40]$ instead of the four discrete rates closes the hole
equally ($38.95$~dB at $6\times$) while leaving the trained rates within $0.06$~dB, and every combination of a
plain or low-frequency encoding with either rate distribution lands on the same curve ($38.94$ to $38.96$ at
$6\times$): the hole is an artefact of the one combination of high-frequency features with a discrete rate set,
not of the hypernetwork idea. We use the plain-MLP hypernetwork as the final
configuration; the models reported above were trained with the Fourier encoding and are unchanged at the trained
rates. A second seed of the plain-MLP brain model reproduces its curve within $0.04$~dB at every rate, trained or
unseen ($41.37/38.95/37.31/35.38/34.11/32.75/31.78$ vs.\ $41.33/38.94/37.31/35.40/34.12/32.77/31.79$~dB).
Retrained with the plain MLP, the $120^3$ brain model is within $0.05$~dB of its Fourier-encoded twin at
the trained rates and gives $43.11/40.74/39.34/37.59/36.50/34.99/33.91$~dB at $4/6/8/12/16/24/32\times$; the knee
model is within $0.06$~dB of its twin at the trained rates and traces a smooth curve through the unseen ones: $37.88/34.73/33.04/31.20/30.15/28.80/27.95$~dB
at $4/6/8/12/16/24/32\times$ (CG-5, true 30~dB). (ii) Modulating the
weights instead of the activations (HyperLowRank, rank-4 kernel deltas on every $3^3$ convolution) is
$0.3$ to $0.4$~dB \emph{worse} than FiLM at the same training budget ($40.92/36.96/33.78$ vs $41.33/37.31/34.16$)
and costs more; FiLM is the right form of hypernetwork here.

\begin{figure}[t]
\centering
\includegraphics[width=0.9\textwidth]{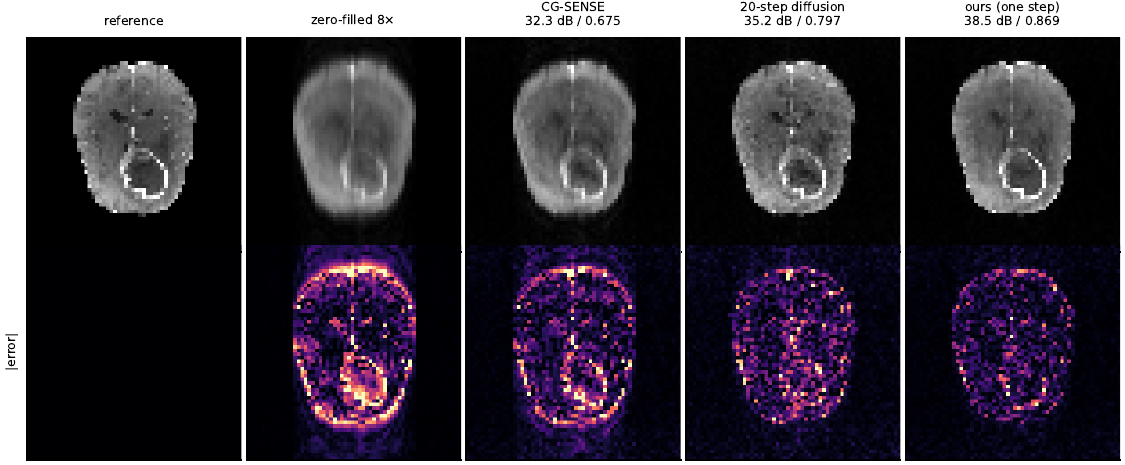}
\caption{One volume of the BraTS $60^3$ set at $8\times$ (30~dB), with $|$error$|$ underneath on a common
scale. One measurement-conditioned network evaluation plus a warm CG projection removes residual aliasing that
the 20-step diffusion prior leaves behind, at $1/20$ of the network cost.}
\label{fig:qual}
\end{figure}

\subsection{Supervised baselines, resolution transfer, knee, the sharpness dial and mixed anatomy}
\paragraph{Supervised baselines and the cost-quality front.} Every row below uses our backbone, data, masks,
noise realisations, data-consistency solver and metrics; the only differences are the method and the number of
network evaluations (NFE). The unrolled rows are our own recipe with $K>1$ iterations of
[network $\to$ warm CG], i.e.\ MoDL/VarNet-style supervised reconstructors initialised from the same one-step
prior. PSNR / Wang-SSIM, 10 volumes, true 30~dB.
\begin{center}\footnotesize
\adjustbox{max width=\textwidth}{%
\begin{tabular}{lcccc}
\toprule
 & NFE & $4\times$ & $8\times$ & $16\times$ \\
\midrule
CG-SENSE (no prior)                              & 0  & 34.80 / 0.678 & 32.04 / 0.631 & 30.31 / 0.611 \\
Unconditional one-step prior + CG                & 1  & 32.96 / 0.699 & 30.96 / 0.656 & 29.59 / 0.627 \\
Unconditional one-step prior + CG                & 4  & 35.40 / 0.664 & 32.50 / 0.601 & 30.52 / 0.559 \\
\quad the same prior, corrected training settings & 1 & 35.08 & 31.92 & 30.09 \\
\quad the same prior, corrected training settings & 4 & 38.80 & 34.36 & 31.42 \\
EDM patch-diffusion prior + CG                   & 20 & 39.83 / 0.845 & 34.22 / 0.762 & 31.04 / 0.727 \\
EDM patch-diffusion prior + CG                   & 80 & 39.95 / 0.847 & 34.20 / 0.761 & 30.93 / 0.724 \\
\midrule
Ours, one step + hyper$(\lambda,R)$              & 1  & 41.33 / 0.876 & 37.31 / 0.858 & 34.12 / 0.840 \\
Ours, unrolled $3\times$ + hyper$(\lambda,R)$    & 3  & 43.75 / 0.913 & 38.62 / 0.879 & 34.78 / 0.850 \\
Ours, unrolled $5\times$ + hyper$(\lambda)$ (HyperRecon form) & 5 & 44.72 / \textbf{0.956} & 39.54 / 0.910 & 35.41 / 0.851 \\
Ours, unrolled $5\times$, fixed $\lambda$ (MoDL/VarNet form)  & 5 & 45.33 / 0.944 & 39.79 / 0.923 & 35.56 / 0.891 \\
Ours, unrolled $5\times$ + hyper$(\lambda,R)$    & 5  & \textbf{46.01} / 0.947 & \textbf{40.13 / 0.926} & \textbf{35.60 / 0.892} \\
\bottomrule
\end{tabular}}
\end{center}
Three readings. (i) \emph{The front, not the point.} One evaluation is the cheapest useful operating point, not
the best one: three and five evaluations buy a further $+2.4/+1.3/+0.7$ and $+4.7/+2.8/+1.5$~dB. Every point of
the front is above the 20- and 80-step diffusion prior, and the five-evaluation point beats it by
$+6.2/+5.9/+4.6$~dB with four times fewer network evaluations. Across the seven rates we evaluate
($4$ to $32\times$, three of them never trained on) the front reads $41.3/38.9/37.3/35.4/34.1/32.8/31.8$ at one
evaluation and $46.0/42.5/40.1/37.6/35.6/34.1/32.8$ at five. (ii) \emph{This is a negative result for
plug-and-play with a generative prior in the paired-data regime}, which is the regime this line of work
usually reports in: a supervised reconstructor with the same backbone, trained through the same solver, is far
ahead at lower cost. The generative prior keeps the advantage it was designed for, needing no paired data, but that advantage should be stated, not assumed. (iii) \emph{Amortising the wrong hyperparameter costs
accuracy.} The unrolled model with a hypernetwork over $\lambda$ (the HyperRecon setting) is $0.6/0.25/0.15$~dB
\emph{below} the same model trained at a fixed $\lambda$; the $(\lambda,R)$ hypernetwork of the one-step model,
by contrast, is $0.15$ to $0.30$~dB above its fixed-$\lambda$ counterpart at every rate (Sec.~\ref{sec:results}).
The difference is what the hyperparameter does: with a warm-started solver $\lambda$ barely changes the
solution, while $R$ changes it entirely. The same holds at the strong end of the front: with five evaluations,
conditioning on $(\lambda,R)$ is worth $+0.33$~dB at $8\times$ over the fixed-$\lambda$ model ($10/10$ volumes,
interval excluding zero) and $+0.41$~dB at the unseen $6\times$, so the amortisation is not a crutch for a weak
model.

\paragraph{Resolution transfer: BraTS-24 $120^3$ evaluation set (10 volumes, 8-coil, 30~dB).}
\begin{center}\footnotesize
\adjustbox{max width=\textwidth}{%
\begin{tabular}{llcccc}
\toprule
 & trained at & NFE & $4\times$ & $8\times$ & $16\times$ \\
\midrule
CG-SENSE                                   & none    & 0  & 35.29 / 0.601 & 31.53 / 0.499 & 29.74 / 0.481 \\
Unconditional one-step flow prior + CG     & $120^3$ & 1  & 35.83 / 0.603 & 31.83 / 0.544 & 29.82 / 0.517 \\
Unconditional one-step flow prior + CG     & $120^3$ & 4  & 37.84 / 0.653 & 34.50 / 0.567 & 31.53 / 0.518 \\
Unconditional one-step flow prior + CG, zero-shot & $60^3$ & 4 & 37.32 / 0.624 & 33.70 / 0.518 & 30.79 / 0.454 \\
EDM patch prior + CG (20 steps)            & $120^3$ & 20 & 41.73 / 0.868 & 36.03 / 0.686 & 32.03 / 0.623 \\
\midrule
Ours, hypernetwork $(\lambda,R)$, \textbf{zero-shot} & $60^3$ & 1 & 41.67 / 0.840 & \textbf{36.95 / 0.758} & \textbf{33.93 / 0.738} \\
Ours, fixed-$\lambda$ specialist, native & $120^3$ & 1 & 42.99 / 0.899 & 39.16 / 0.861 & 36.34 / 0.851 \\
Ours, hypernetwork $(\lambda,R)$, native (CG-5 / CG-20 identical to $0.01$~dB) & $120^3$ & 1 & \textbf{43.15 / 0.902} & \textbf{39.39 / 0.863} & \textbf{36.51 / 0.853} \\
\bottomrule
\end{tabular}}
\end{center}
Trained natively at $120^3$ (10k steps, 4~h on one Titan RTX) the one-step model is $+1.4/+3.4/+4.5$~dB above
the 20-step diffusion prior at $4/8/16\times$, again with the margin growing with the acceleration (a second
training seed gives $43.10/39.39/36.50$~dB). The hypernetwork model again beats its fixed-$\lambda$ specialist
at every rate, $+0.15/+0.23/+0.17$~dB, $10/10$ volumes each (95\% CIs $[+0.09,+0.21]$, $[+0.12,+0.34]$,
$[+0.10,+0.24]$). The
$60^3$-trained model applied unchanged to $120^3$ volumes (same $32^3$ patch network, the thumbnail simply
pools a larger volume) ties the natively-trained 20-step diffusion prior at $4\times$ and is $+0.9/+1.9$~dB above
it at $8/16\times$, i.e.\ $1.5$ to $2.6$~dB below native training (under the light-noise convention, CG-20:
$42.53/37.52/34.30$ vs.\ EDM-20 $43.36/37.45/32.88$). Training the $60^3$ model three times longer does not change
its transfer ($41.73/36.86/33.95$): the gap to native training is a resolution effect, not a budget effect. Implicit-kernel convolutions (hyper-convolutions \citep{ma2023hyper}), whose taps
can be re-sampled at the new resolution, do \emph{not} help: on the conditioned reconstructor the
implicit-kernel network is $1.5$~dB worse natively ($39.80/35.98/33.25$), its zero-shot transfer with the
$3^3$ kernel re-sampled at 3 taps is $2.2$~dB below the dense network's plain transfer
($39.45/34.94/32.00$), and re-sampling at 5 taps is catastrophic ($29.76/24.37/22.10$); the unconditional
prior shows the same pattern. Resolution transfer here comes from the patch/thumbnail parametrisation, not
from the convolution kernels.

\paragraph{SKM-TEA knee, quarter resolution $128{\times}128{\times}80$ (10 volumes, 30~dB).}
\begin{center}\footnotesize
\adjustbox{max width=\textwidth}{%
\begin{tabular}{lcccc}
\toprule
 & NFE & $4\times$ & $8\times$ & $16\times$ \\
\midrule
CG-SENSE (prior-free)                       & 0  & 38.04 / 0.916 & 30.34 / 0.737 & 26.95 / 0.597 \\
Unconditional one-step flow prior + CG      & 4  & 38.50 / 0.935 & 31.62 / 0.792 & 27.46 / 0.653 \\
EDM patch prior + CG (20 steps), re-trained in our pipeline & 20 & \textbf{38.90} / 0.932 & 31.55 / 0.757 & 27.35 / 0.590 \\
Ours, hypernetwork $(\lambda,R)$, 10k steps, CG-5 & 1 & 37.94 / \textbf{0.937} & \textbf{33.10 / 0.852} & \textbf{30.16 / 0.782} \\
Ours, fixed $\lambda{=}0.03$ specialist, 10k steps, CG-20 & 1 & 39.76 / 0.946 & 33.53 / 0.834 & 30.22 / 0.751 \\
Ours, hypernetwork $(\lambda,R)$, 10k steps, CG-20 & 1 & 39.74 / 0.945 & 33.47 / 0.827 & 30.13 / 0.732 \\
Ours, unrolled $3\times$ + hypernetwork $(\lambda,R)$, CG-20 & 3 & \textbf{40.86 / 0.965} & \textbf{35.60 / 0.915} & \textbf{31.99 / 0.827} \\
\bottomrule
\end{tabular}}
\end{center}
Knee at $4\times$ is the solver-bound regime the family's earlier work identified: the prior-free CG-SENSE is
already at $38.0$~dB and the 20-step prior adds $0.9$~dB. There the one-step model with its \emph{five}-iteration
projection is $0.1$~dB below CG-SENSE; its best $\lambda$ sits at the bottom of the grid, i.e.\ it wants more
data consistency than five warm iterations give. Twenty warm iterations (still one network pass) fix this:
$39.74/33.47/30.13$~dB, i.e.\ $+0.8$~dB above the 20-step prior at $4\times$ as well as $+1.9/+2.8$~dB at
$8/16\times$. Under the light-noise convention the knee numbers become CG-SENSE $41.05/31.85/27.57$ and our EDM-20 prior
$42.12/33.89/28.50$~dB, and the one-step model with twenty
warm iterations reaches $42.15/34.97/30.87$~dB (SSIM $0.972/0.887/0.794$): level with the 20-step prior at
$4\times$, above the published 80-step number, and $+1.1/+2.4$~dB above the 20-step prior at $8/16\times$. On the brain
sets the number of warm iterations is immaterial (five vs.\ twenty: within $0.1$~dB under the light-noise convention,
$0.05$ to $0.5$~dB in favour of five under the true 30~dB, where the extra iterations fit noise unless $\lambda$
is raised); on knee at $4\times$, where the data alone give $41$~dB, they are worth $+1.8$~dB. The test-time
projection is therefore reported as: warm-started CG, 20 iterations, $\lambda$ from the grid
$\{10^{-4},\dots,0.3\}$ per rate .
The plain-MLP retrain of the knee model reproduces these numbers ($39.74/33.59/30.28$ true 30~dB and
$42.11/34.92/30.86$ under the light-noise convention, CG-20), and training three times longer adds $0.1$ to $0.4$~dB
($39.88/33.85/30.52$ and $42.35/35.21/31.11$): under the light-noise convention the one-step knee model at $4\times$
is $0.6$~dB above the published 80-step number. The front holds on knee as well: three evaluations of the same recipe give $40.86/35.60/31.99$~dB, $+1.17/+2.02/+1.77$~dB over one evaluation on every one of the 10 volumes (paired intervals $[+1.1,+1.3]$, $[+1.9,+2.1]$, $[+1.6,+1.9]$) and $+2.0/+4.1/+4.6$~dB over the 20-step diffusion prior at a seventh of its network evaluations.

\paragraph{A dial that changes the answer: a negative result .} A hyperparameter is only worth amortising
if moving it moves the reconstruction. We tried to build a user-facing sharpness dial by interpolating the
training loss between image-domain and gradient-domain fidelity, $\alpha\in[0,1]$, and exposing $\alpha$ to the
hypernetwork. One model swept across $\alpha$ moves by $0.02$~dB and $0.002$ SSIM at $4\times$
($42.32\to42.34$~dB, SSIM $0.888\to0.887$ from $\alpha{=}0$ to $1$); three models each trained at a single
$\alpha$ span $0.15$~dB. Attributing what little there is, on 10 volumes evaluated at $\alpha{=}0$ (paired,
against the same model trained with a plain MSE loss): sampling the mixture and telling the network gives
$+0.07/+0.07/+0.10$~dB at $4/8/16\times$, sampling it without telling the network gives $+0.09/+0.06/+0.07$,
and giving the network the input without ever varying the loss gives $+0.02/+0.01/+0.01$. The small gain is a
loss-augmentation effect, not conditioning, and none of it is a dial a radiologist could use. We report this
because the failure is informative: HyperRecon's promise of browsing a family of reconstructions needs a
hyperparameter whose effect survives the data-consistency step, and a gradient-domain reweighting of the
training loss does not.

\paragraph{One model across anatomies .} Training the same hypernetwork reconstructor on brain ($60^3$)
and knee (QR) volumes jointly (datasets sampled uniformly, 20k steps so that each sees the 10k of the single-dataset models)
costs nothing on knee ($39.75/33.56/30.23$ vs.\ $39.74/33.59/30.28$~dB for the knee-only
model, CG-20) and $0.8$~dB on brain ($40.49/36.52/33.42$ vs.\ $41.33/37.31/34.12$, CG-5) at every rate; the
model must infer the anatomy from the zero-filled image. Neither remedy closes the gap: an anatomy code as a
third hypernetwork input recovers $0.1$~dB ($40.60/36.63/33.52$) and a 40k-step budget recovers $0.3$~dB
($40.77/36.77/33.57$), so the cost is capacity or optimisation, not missing conditioning.

\subsection{Learned acquisition in full}
\paragraph{The acquisition as an operating point .} The protocol-matched tables keep the protocol's sampling
pattern. If the model may also choose \emph{where} to sample, the hypernetwork can emit the Cartesian mask for
the requested acceleration (Sec.~\ref{sec:acq}), trained jointly with the reconstructor on
unmasked k-space. Against an identical model trained and evaluated with the protocol's variable-density
Gaussian mask (same data, same 150-volume validation set, same 10 evaluation volumes, CG-5):
\begin{center}\footnotesize
\adjustbox{max width=\textwidth}{%
\begin{tabular}{lcccc}
\toprule
acquisition & $4\times$ & $8\times$ & $16\times$ & $32\times$ \\
\midrule
protocol mask (variable-density Gaussian)          & 40.11 / 0.842 & 36.28 / 0.813 & \textbf{33.52 / 0.835} & \textbf{31.63 / 0.831} \\
learned mask, model evaluated with the protocol mask & 38.50 / 0.792 & 34.94 / 0.752 & 31.25 / 0.675 & 29.00 / 0.639 \\
\textbf{learned mask}, model evaluated with its own   & \textbf{44.56 / 0.925} & \textbf{38.20 / 0.852} & 32.89 / 0.800 & 30.46 / 0.782 \\
\bottomrule
\end{tabular}}
\end{center}
\paragraph{The same experiment on clinical knee data.} SKM-TEA ships fully sampled acquisitions with
scanner-derived coil maps, so the experiment can be repeated on real anatomy and real sensitivities rather than
simulated ones: same recipe, same protocol-mask control, 33-volume validation set.
\begin{center}\footnotesize
\begin{tabular}{lcccc}
\toprule
acquisition (SKM-TEA knee, real raw data, 10 volumes) & $4\times$ & $8\times$ & $16\times$ & $32\times$ \\
\midrule
protocol mask                 & 40.07 & 33.89 & 30.53 & 28.29 \\
\textbf{learned mask}         & \textbf{42.45} & \textbf{35.52} & \textbf{31.41} & \textbf{28.95} \\
\midrule
paired gain                   & $+2.34$ & $+1.57$ & $+0.86$ & $+0.67$ \\
volumes won                   & 10/10 & 10/10 & 10/10 & 10/10 \\
\bottomrule
\end{tabular}
\end{center}
Both models are trained for 30k steps; at 10k the same pair reads $+2.06/+1.19/+0.81/+0.60$, so training both
sides longer widens the gap rather than closing it. On real data the learned acquisition wins at \emph{every}
rate, including the $16$ and $32\times$ where the simulated brain experiment crossed over. The crossover is therefore a property of that simulation (small volumes whose information is concentrated in a few central lines) and not of
the method.

\begin{figure}[t]
\centering
\includegraphics[width=0.6\textwidth]{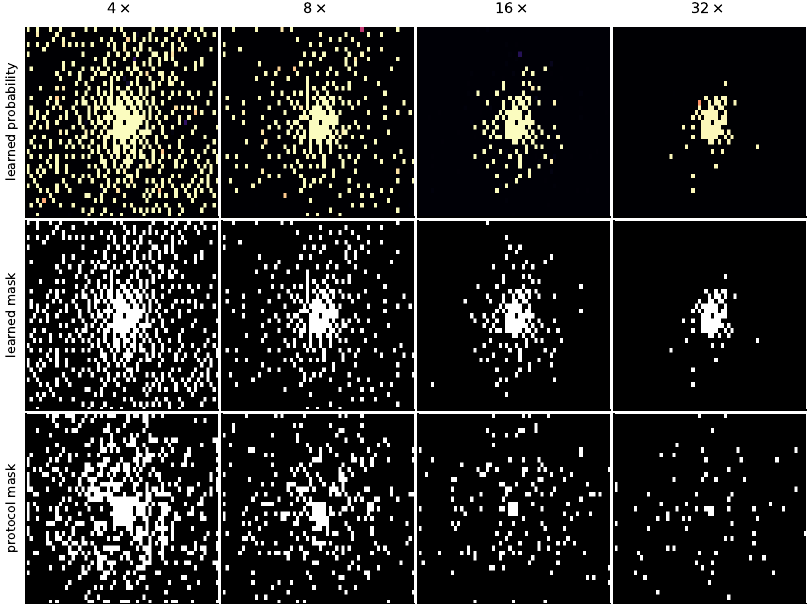}
\caption{What the model chooses to acquire (BraTS $60^3$). Top: the probability the hypernetwork emits for the
requested acceleration; middle: the phase-encode locations actually sampled (top-$YZ/R$); bottom: the
protocol's variable-density Gaussian mask at the same sampling fraction. At a matched budget the learned
pattern concentrates far more strongly on the centre (the fraction of the central $12\times12$ block it keeps is
$0.69/0.59/0.50/0.44$ at $4/8/16/32\times$ against $0.59/0.38/0.20/0.07$ for the protocol) and its
probabilities saturate near $0$ and $1$, i.e.\ it makes decisions rather than jitter.}
\label{fig:masks_app}
\end{figure}

A second seed of the protocol-mask control agrees with the first to $0.02$ to $0.03$~dB at every rate, so
the differences below are two orders of magnitude above the training noise. The comparison is also not against
an untuned baseline: retraining the same model with the Gaussian widened ($22\times15$) or narrowed
($10\times7$) moves it by at most $0.6$~dB and never past the learned mask
($40.11/36.28/33.52/31.63$ for the protocol widths, $39.52/36.33/33.78/31.74$ wider, $38.64/35.01/32.44/31.19$
narrower). The narrower mask, the hand-made way to ``concentrate on the centre'', is the \emph{worst} of the
three, so what the learned pattern buys is its shape, not simply more central density. Over four seeds (two of them trained on a different machine and from a different data path, with the coils and
the full k-space regenerated on the fly from the targets rather than read from a stored measurement LMDB) the
pair reads $44.59\pm0.04$ / $38.37\pm0.11$ / $33.25\pm0.35$ /
$30.75\pm0.21$~dB for the learned mask against $40.13\pm0.02$ / $36.32\pm0.03$ / $33.50\pm0.03$ /
$31.58\pm0.07$ for the protocol mask. The seed-to-seed spread is $0.02$ to $0.35$~dB against gains of $4.46$ and
$2.05$~dB at $4$ and $8\times$, and the small deficits at $16$ and $32\times$ ($-0.26$, $-0.83$) are equally
stable. Learning the acquisition is worth
$+4.5$~dB at $4\times$ and $+1.9$~dB at $8\times$, an order of magnitude more than any other operating point we
amortise, and it \emph{costs} $0.6$ to $1.2$~dB at $16$ and $32\times$. The
middle row shows the two halves are bound together: the same network loses $1.6$ to $6$~dB when its own mask is
replaced by the protocol's. Fig.~\ref{fig:masks_app} explains the crossover: the learned pattern spends its budget on the centre, which is the
right trade at $4$ to $8\times$ and the wrong one when only a few percent of the plane can be sampled and
resolution has to come from somewhere. Two controls locate the cause. It is \emph{not} that one probability map
is stretched over $4$ to $32\times$: a sampler trained for a single rate is no better than the shared one
($44.22$ vs.\ $44.56$~dB at $4\times$; $32.55$ vs.\ $32.89$ at $16\times$), so sharing rates in fact regularises
the sampler. And a sampler trained \emph{only} at $16\times$ still loses to the protocol mask by $0.97$~dB
($0/10$ volumes, interval excluding zero), so at that budget centre-heavy sampling is simply the wrong trade.
The crossover is therefore a property of the sampling budget, and it moves with the amount of data per volume. At $120^3$, where
each volume carries eight times more data, the same experiment with a matched control gives
\begin{center}\footnotesize
\adjustbox{max width=\textwidth}{%
\begin{tabular}{lcccc}
\toprule
acquisition ($120^3$, 10 volumes) & $4\times$ & $8\times$ & $16\times$ & $32\times$ \\
\midrule
protocol mask            & 43.12 / 0.902 & 39.34 / 0.864 & 36.49 / 0.854 & 33.90 / 0.870 \\
\textbf{learned mask}    & \textbf{45.82 / 0.925} & \textbf{41.04 / 0.902} & \textbf{37.06 / 0.880} & \textbf{34.05} / 0.854 \\
\midrule
paired gain              & $+2.70$ & $+1.69$ & $+0.57$ & $+0.16$ \\
volumes won / interval   & 10/10, $[+2.5,+2.9]$ & 10/10, $[+1.5,+1.9]$ & 9/10, $[+0.2,+0.9]$ & 7/10, $[-0.1,+0.4]$ \\
\bottomrule
\end{tabular}}
\end{center}
so the learned acquisition wins at every rate, by $2.7$~dB where the budget is comfortable and by nothing once
only $3\%$ of the plane may be sampled. The gain is not a short-training artefact either: training the learned
sampler for 30k steps instead of 10k at $60^3$ adds a further $+0.51/+0.76/+0.36/+0.14$~dB
($45.07/38.96/33.25/30.60$); its matched control, the protocol-mask model trained 30k steps, gives
$40.59/36.78/33.94/31.96$, so the paired gain is unchanged ($+4.47/+2.17/-0.70/-1.36$~dB, $10/10$ at $4$ and
$8\times$, $0/10$ at $16$ and $32\times$) and neither side was stopped early relative to the other. Designing the acquisition is thus the single most valuable thing the
hypernetwork does: it is worth an order of magnitude more than conditioning on $\lambda$ (nothing), on the loss
trade-off (nothing) or on the acceleration ($0.2$~dB), and it is available at no extra test-time cost, since
the mask is produced by the same forward pass that configures the reconstructor. Note that this row changes the acquisition and therefore
leaves the protocol of the tables above; it is reported separately from the protocol-matched tables above, and only
on BraTS, where we simulate the k-space and can store it unmasked.

\paragraph{Without fully-sampled targets: self-supervision costs almost nothing .} Every result so far
assumes paired data, which a clinical archive does not have. We therefore train the same reconstructor with no
target at all: the acquired samples are split, the model sees one half, and the loss is how well it predicts
the other half in k-space (the k-space centre stays in the visible half, without which the coil sensitivities
and the low frequencies are unobservable). On the \emph{real} raw knee data of SKM-TEA (33-volume validation
set, 10 evaluation volumes, CG-20):
\begin{center}\footnotesize
\begin{tabular}{lcccc}
\toprule
supervision & $4\times$ & $8\times$ & $16\times$ & $32\times$ \\
\midrule
fully-sampled targets                    & 39.92 & \textbf{33.64} & \textbf{30.29} & \textbf{28.06} \\
\textbf{none} (k-space split only)       & \textbf{39.93} & 33.41 & 29.99 & 27.67 \\
\midrule
paired difference                        & $-0.03$ & $-0.23$ & $-0.31$ & $-0.40$ \\
volumes won / interval                   & 5/10, $[-0.09,+0.03]$ & 0/10 & 0/10 & 0/10 \\
\bottomrule
\end{tabular}
\end{center}
At $4\times$ the two are indistinguishable, and the gap grows to $0.4$~dB only at $32\times$, where the held-out
half of a very sparse acquisition is itself a poor teacher. Both are far above CG-SENSE ($38.0/30.3/27.0$) and
the 20-step diffusion prior ($38.9/31.6/27.4$) on the same data. The practical reading is that the fully-sampled
archive our supervised numbers rely on is not required: the scan supervises itself, and the amortised operating
point rides along unchanged.

\paragraph{Where the learned acquisition sits among untrained and trained baselines.} On the same
validation set and noise level, with the sparsity weight or $\lambda$ swept per rate (10 volumes):
\begin{center}\footnotesize
\begin{tabular}{lcccc}
\toprule
BraTS-60, 30~dB & $4\times$ & $8\times$ & $16\times$ & $32\times$ \\
\midrule
CG-SENSE, no prior (no training)          & 32.35 & 29.64 & 27.92 & 26.52 \\
$\ell_1$-wavelet CS, FISTA (no training)  & 36.21 & 30.23 & 27.49 & 26.05 \\
ours, protocol mask                       & 40.11 & 36.28 & \textbf{33.52} & \textbf{31.63} \\
\textbf{ours, learned mask}               & \textbf{44.56} & \textbf{38.20} & 32.89 & 30.46 \\
\bottomrule
\end{tabular}
\end{center}
Classical compressed sensing sits where one expects: well above the prior-free solver at $4\times$, level with
it by $16\times$, and $4$ to $8$~dB below the trained reconstructors throughout. The learned acquisition adds more
on top of our model at $4\times$ ($+4.45$~dB) than classical sparsity adds on top of CG-SENSE ($+3.86$~dB).

\paragraph{Is this just LOUPE in 3D? The reconstructor decides.} LOUPE \citep{bahadir2019learning} learns the sampling
pattern jointly with a \emph{feed-forward} reconstructor: the network maps the zero-filled image to the
reconstruction with no data-consistency step. Ours learns it through a warm-started CG projection. Training all
four combinations at $60^3$, everything else identical (10 volumes):
\begin{center}\footnotesize
\begin{tabular}{lcccc}
\toprule
reconstructor / acquisition & $4\times$ & $8\times$ & $16\times$ & $32\times$ \\
\midrule
feed-forward, protocol mask            & 37.19 & 35.09 & 32.97 & 31.33 \\
feed-forward, learned mask (LOUPE form) & 35.57 & 35.86 & 32.11 & 28.61 \\
warm CG, protocol mask                 & 40.11 & 36.28 & \textbf{33.52} & \textbf{31.63} \\
\textbf{warm CG, learned mask (ours)}  & \textbf{44.56} & \textbf{38.20} & 32.89 & 30.46 \\
\midrule
acquisition gain, feed-forward         & $-1.61$ & $+0.77$ & $-0.86$ & $-2.71$ \\
acquisition gain, warm CG              & $+4.48$ & $+1.93$ & $-0.64$ & $-1.16$ \\
\bottomrule
\end{tabular}
\end{center}
With a feed-forward reconstructor, learning the acquisition \emph{hurts} at $4\times$ ($-1.61$~dB); with the
data-consistency projection the same learned acquisition is worth $+4.48$~dB. The feed-forward model also fails
to improve as the budget grows (it scores \emph{lower} at $4\times$ than at $8\times$), which is the same
fact seen from the other side: a network that never enforces the measurement cannot cash in extra samples, so a
better sampling pattern buys it nothing. The two halves are therefore not separable contributions. What we add
to LOUPE is not a bigger mask network; it is the observation that learned acquisition and enforced data
consistency only pay together, and the $3$D, multi-coil, acceleration-conditioned setting in which that can be
measured.

\paragraph{The acquisition gain is not the hypernetwork's.} Every acquisition experiment above uses the
$(\lambda,R)$ hypernetwork, so the two contributions could be confounded. Repeating the pair with no
hypernetwork at all (a single fixed $\lambda$, no conditioning) separates them: the learned mask gains
$+4.37/+2.06/+0.09/-0.56$~dB at $4/8/16/32\times$ against its own protocol-mask control
($44.43/38.35/33.55/30.92$ vs.\ $40.07/36.30/33.46/31.48$; $10/10$ volumes at $4$ and $8\times$), essentially
the same curve as the $+4.45/+1.92/-0.63/-1.17$ measured with the hypernetwork. The two are orthogonal: the
hypernetwork buys coverage of the whole acceleration range for a few tenths of a decibel, the learned
acquisition buys information for two to four decibels, and either can be adopted without the other.

\paragraph{The acquisition gain grows with a stronger reconstructor.} The natural objection to a learned
acquisition is that it only compensates for a weak reconstructor and will evaporate once the reconstructor
improves. The opposite happens. Training the same pair with three evaluations instead of one (BraTS-60, 10
volumes):
\begin{center}\footnotesize
\begin{tabular}{lcccc}
\toprule
 & $4\times$ & $8\times$ & $16\times$ & $32\times$ \\
\midrule
one evaluation, protocol mask   & 40.11 & 36.28 & 33.52 & 31.63 \\
one evaluation, learned mask    & 44.56 & 38.20 & 32.89 & 30.46 \\
three evaluations, protocol mask & 41.11 & 36.58 & 33.38 & 31.18 \\
three evaluations, learned mask  & \textbf{47.87} & \textbf{41.78} & \textbf{35.91} & \textbf{31.96} \\
\midrule
acquisition gain, one evaluation    & $+4.45$ & $+1.92$ & $-0.63$ & $-1.17$ \\
acquisition gain, three evaluations & $+6.76$ & $+5.21$ & $+2.53$ & $+0.79$ \\
\bottomrule
\end{tabular}
\end{center}
The gain roughly doubles at $4$ and $8\times$, and the sign flips at $16$ and $32\times$: with three
evaluations the learned acquisition wins everywhere ($10/10$ volumes at every rate, intervals excluding zero).
A better acquisition and a better reconstructor are complements, not substitutes: the stronger model is the
one that can actually cash in the information a better sampling pattern collects. It also explains the
crossover as a property of the weaker operating point rather than of the acquisition.

\paragraph{Self-supervision and learned acquisition do not compose .} Both ingredients work on their
own, and combining them is the obvious next step: no fully-sampled data \emph{and} an acquisition chosen by the
model. It fails, badly and instructively. At $60^3$:
\begin{center}\footnotesize
\begin{tabular}{lcccc}
\toprule
 & $4\times$ & $8\times$ & $16\times$ & $32\times$ \\
\midrule
supervised, protocol mask       & 40.11 & 36.28 & \textbf{33.52} & \textbf{31.63} \\
supervised, learned mask        & \textbf{44.56} & \textbf{38.20} & 32.89 & 30.46 \\
self-supervised, protocol mask  & 39.75 & 36.11 & 33.32 & 31.41 \\
self-supervised, learned mask   & 32.83 & 31.15 & 29.93 & 29.70 \\
\bottomrule
\end{tabular}
\end{center}
Self-supervision alone costs $0.2$ to $0.4$~dB; the learned acquisition alone gains $4.5/1.9$~dB at $4/8\times$;
together they lose $6.9$~dB against self-supervision with the protocol mask. The learned masks say exactly what
happened. Measuring how often a sampled phase-encode location has sampled neighbours, the self-supervised
learned mask sits at $0.94/0.90/0.70/0.55$ across the four rates against $0.20/0.20/0.27/0.46$ for the
supervised one: it acquires in dense clumps. That is the optimum of the objective it was given (a held-out sample surrounded by acquired samples is nearly free to predict) and it
is close to the worst thing to do for
reconstruction, which needs the budget spread over k-space. When the model both sets the exam and sits it, it
optimises the exam. A self-supervised loss that is to be combined with a learned acquisition therefore needs a
held-out set the sampler cannot influence, which we leave to future work.

\paragraph{Adapting the acquisition to the subject: a negative result with a clear cause .} If a mask
per acceleration helps, a mask per \emph{subject} should help more. We split the acquisition in two: every
subject is first acquired with a fixed low-frequency block ($1/32$ of the budget), a fully convolutional policy
reads the energy that block measured at each phase-encode location together with the operating point, and it
chooses the remaining lines; training is end-to-end with the same relaxation. At $120^3$ this is much
\emph{worse} than both references: $35.9/33.7/33.3/31.6$~dB at $4/8/16/32\times$ against $45.8/41.0/37.1/34.1$
for the non-adaptive learned mask and $43.1/39.3/36.5/33.9$ for the protocol mask ($0/10$ volumes at every
rate). The masks say why: the adaptive policy concentrates on the centre even harder than the learned static
mask (the fraction of a central disc it keeps is $0.82/0.74/0.69/0.44$ against $0.71/0.55/0.43/0.31$, and the
central $12\times12$ block saturates at $1.00$). Conditioning on measured energy makes the policy greedy for
high-energy lines, and in k-space the highest-energy lines are the central ones it has already acquired:
energy is not information. The static learned mask never sees the energy and is accountable only to the
reconstruction loss, which is what makes it find the better balance. The failure reproduces across scales, machines and anatomies: at $60^3$ the adaptive policy is
$10.2/6.6/1.9/0.7$~dB below the static learned mask at $4/8/16/32\times$, and on the \emph{real} knee
acquisitions $10.5/5.4/2.6/1.9$~dB below it, losing on every volume in all three settings. We report this because the same trap has been reported for posterior-variance
criteria in active acquisition, and because it locates the design question precisely: an adaptive policy needs
a signal about what is still \emph{unknown}, not about what is large.

\section{Qualitative comparisons}\label{app:qual}
Each figure shows one volume: the same mask, noise draw and $\lambda$ grid for every column, with the two
learned-acquisition columns acquiring their own mask at the same budget. Numbers under the titles are the
displayed volume's PSNR (dB) / SSIM; error maps share one colour scale per figure ($0.2\times$ the reference
maximum).
\IfFileExists{figs/fig_qual_knee_R4.pdf}{\begin{figure}[h]\centering\includegraphics[width=\textwidth]{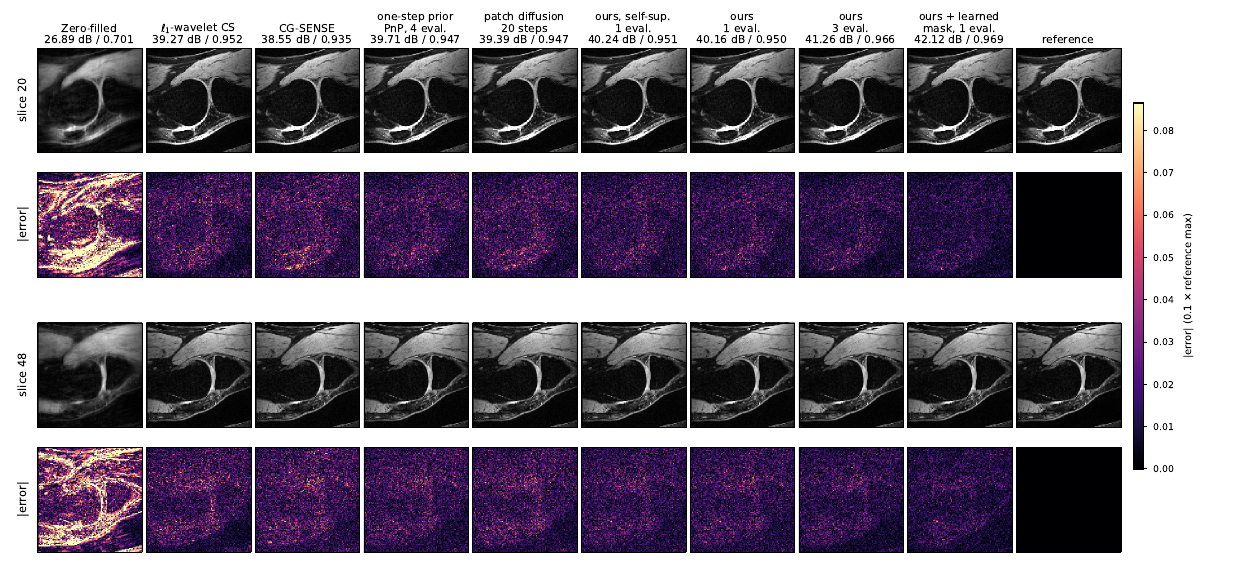}\caption{SKM-TEA knee, $4\times$, volume 0; error scale $0.1\times$ the reference maximum, half that of the other figures, because at this solver-bound rate the residuals are small for every method.}\end{figure}}{}
\IfFileExists{figs/fig_qual_knee_R16.pdf}{\begin{figure}[h]\centering\includegraphics[width=\textwidth]{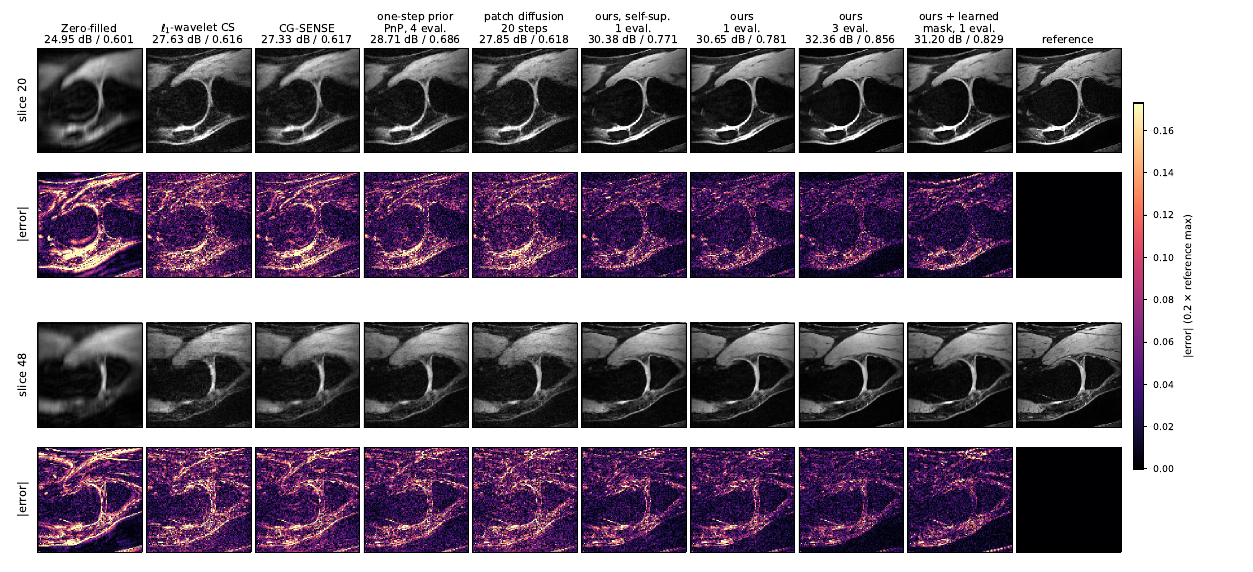}\caption{SKM-TEA knee, $16\times$, volume 0.}\end{figure}}{}
\IfFileExists{figs/fig_qual_brats60_R4.pdf}{\begin{figure}[h]\centering\includegraphics[width=\textwidth]{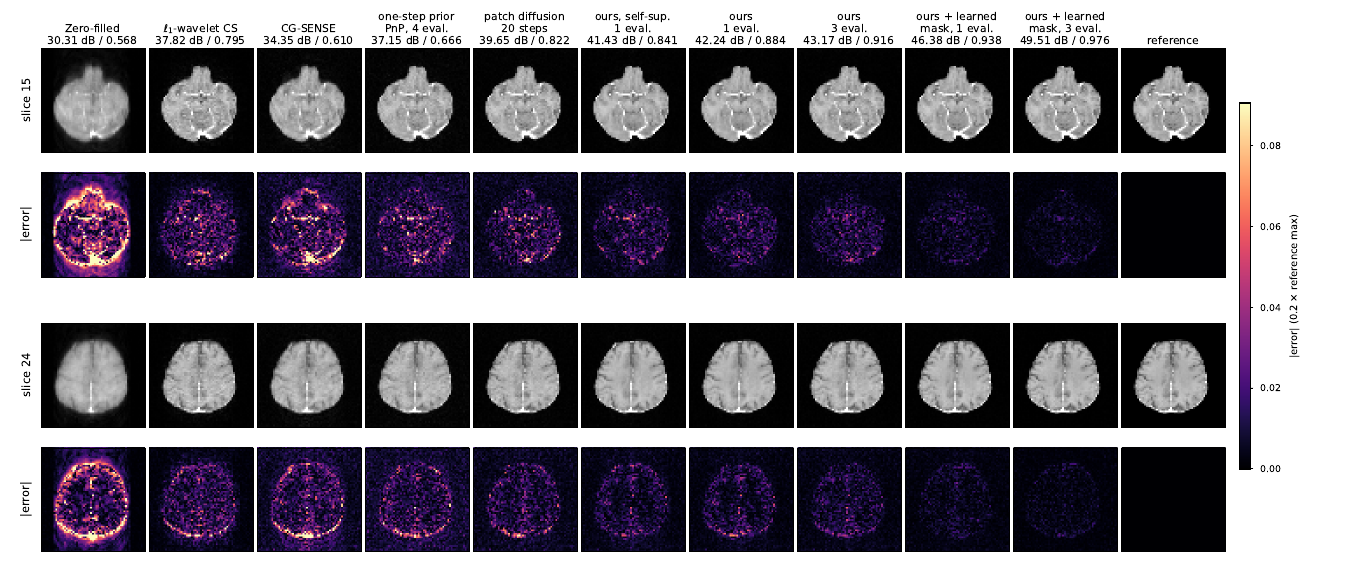}\caption{BraTS $60^3$, $4\times$, volume 3.}\end{figure}}{}
\IfFileExists{figs/fig_qual_knee_R8.pdf}{\IfFileExists{figs/fig_qual_brats60_R8.pdf}{\begin{figure}[h]\centering\includegraphics[width=\textwidth]{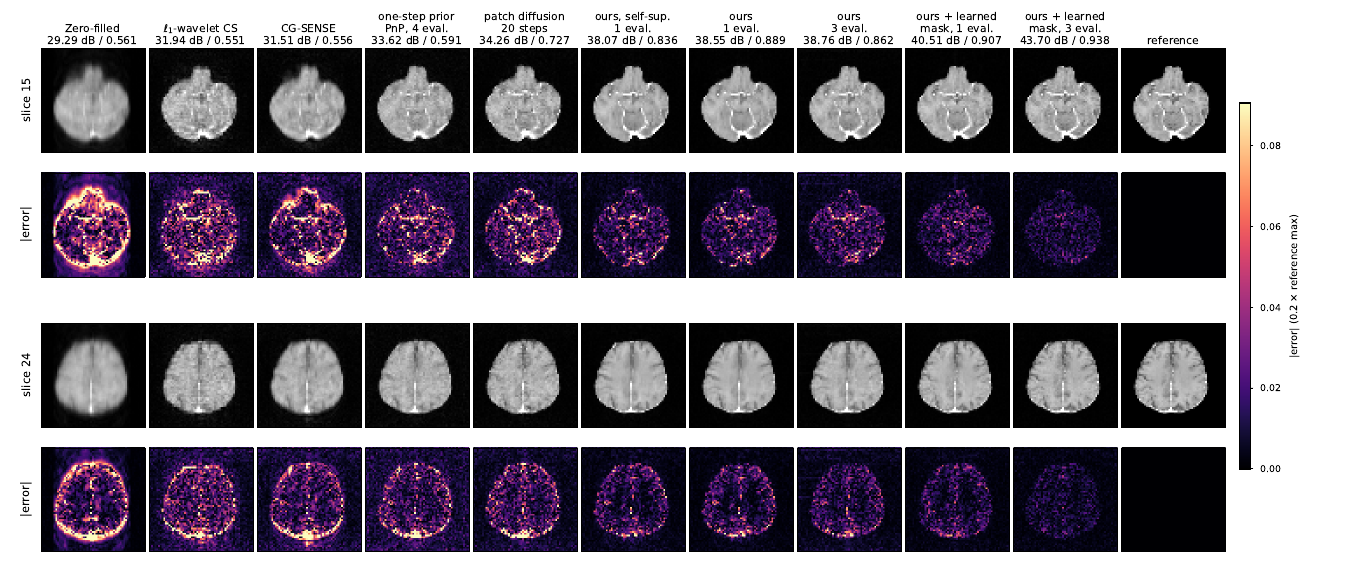}\caption{BraTS $60^3$, $8\times$, volume 3.}\end{figure}}{}}{}
\IfFileExists{figs/fig_qual_brats60_R16.pdf}{\begin{figure}[h]\centering\includegraphics[width=\textwidth]{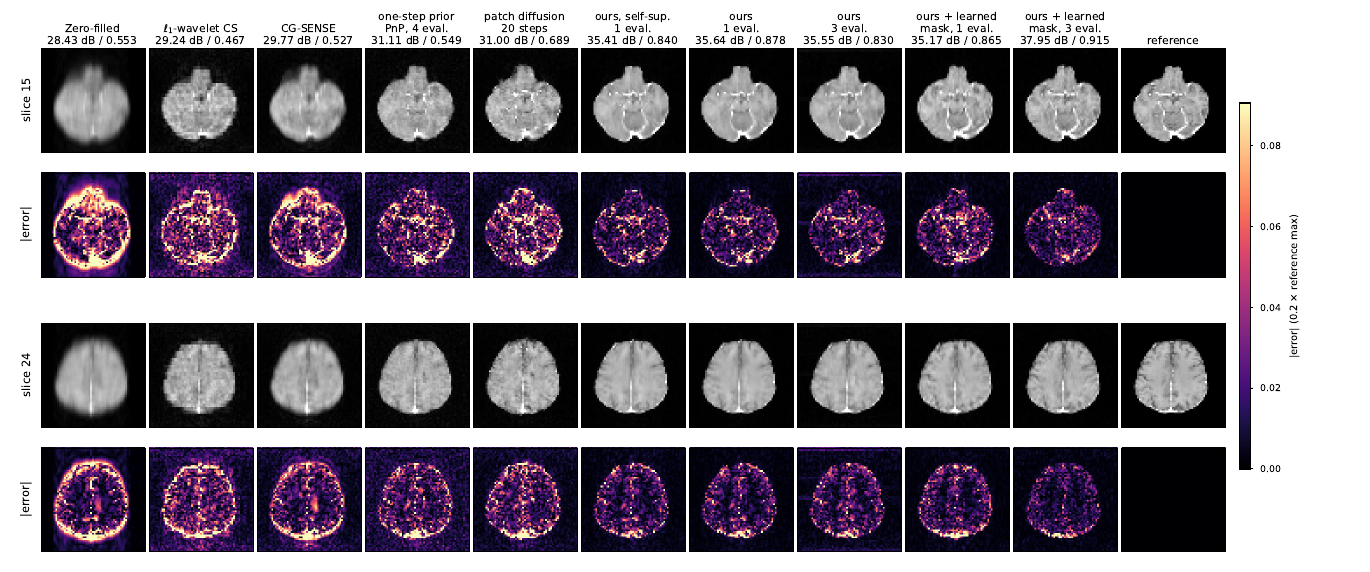}\caption{BraTS $60^3$, $16\times$, volume 3.}\end{figure}}{}
\IfFileExists{figs/fig_qual_brats60_vol0_R8.pdf}{\begin{figure}[h]\centering\includegraphics[width=\textwidth]{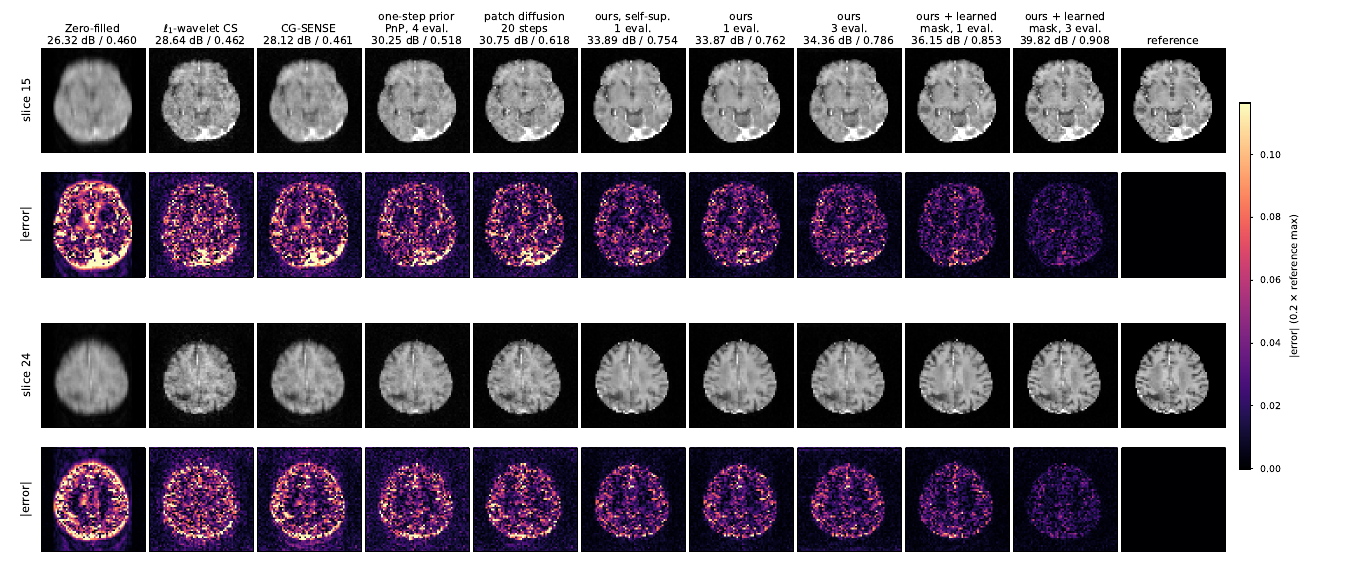}\caption{BraTS $60^3$, $8\times$, volume 0.}\end{figure}}{}
\IfFileExists{figs/fig_qual_brats120_R8.pdf}{\begin{figure}[h]\centering\includegraphics[width=\textwidth]{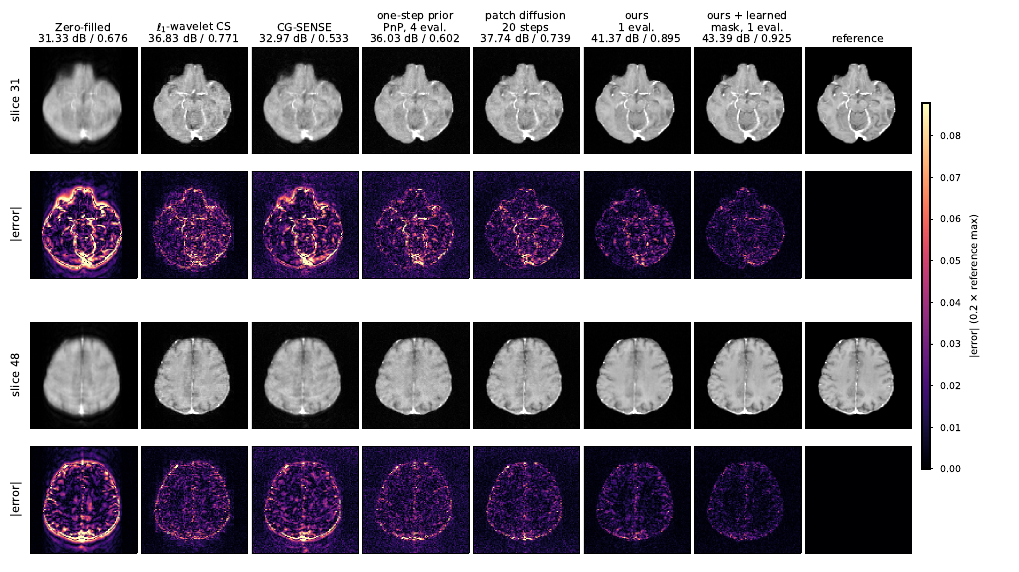}\caption{BraTS $120^3$, $8\times$, volume 3 (no unrolled model was trained at this resolution).}\end{figure}}{}
\IfFileExists{figs/fig_qual_brats120_R4.pdf}{\begin{figure}[h]\centering\includegraphics[width=\textwidth]{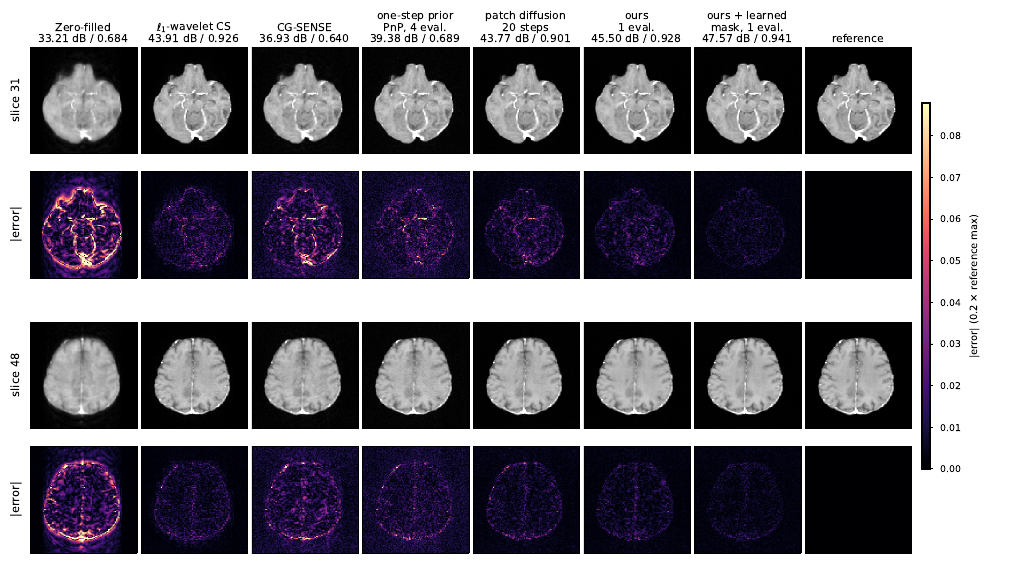}\caption{BraTS $120^3$, $4\times$, volume 3.}\end{figure}}{}

\section{The diffusion prior against its step budget}\label{app:sweep}
\IfFileExists{figs/fig_nfe_sweep.pdf}{\begin{figure}[h]\centering\includegraphics[width=0.95\textwidth]{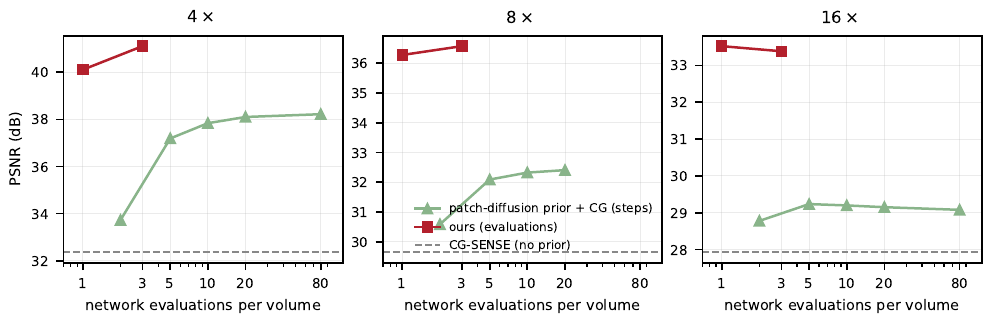}
\caption{PSNR against network evaluations on one split (BraTS $60^3$ full-k-space split, protocol mask, 30~dB, 10
volumes, best $\lambda$ per point): the patch-diffusion prior at $2/5/10/20/80$ sampler steps (each step also runs a
40-iteration CG), our reconstructor at one and three evaluations (each followed by five warm CG iterations), and
the prior-free CG-SENSE.}\label{fig:sweep}\end{figure}}{}
The prior is not under-sampled at 20 steps. On the full-k-space split (Fig.~\ref{fig:sweep}) the 20-step patch-diffusion
prior gives $38.10/32.41/29.15$~dB at $4/8/16\times$ and 80 steps $38.22/32.37/29.08$; going the other way, 10 steps give
$37.84/32.33/29.20$, 5 steps $37.19/32.09/29.23$ and 2 steps $33.72/30.59/28.78$. At $16\times$ the prior is flat from
five steps on; at $4\times$ it gains $0.9$~dB between 5 and 20 steps and $0.1$~dB beyond. On the same split our one
evaluation gives $40.11/36.28/33.52$~dB and three evaluations $41.11/36.58/33.38$: one network pass is above every point of
the diffusion prior's curve, including its converged value, and at matched cost (five network passes on either side) the
gap is $+2.9/+4.2/+4.3$~dB in favour of one pass of ours against five of the prior. The prior-free CG-SENSE is
$32.35/29.64/27.92$~dB here.

\subsection*{Learned acquisition at accelerations never trained on}
\begin{table}[h]\centering\small
\caption{\textbf{The mask head interpolates.} BraTS $60^3$, all 150 validation volumes, 30~dB, PSNR (dB). The hypernetwork was
trained on $R\in\{4,8,16,32\}$; here it is asked for $6/12/24\times$, emitting both the FiLM modulation and the
sampling pattern for a rate it never saw. Paired gain of the learned over the protocol mask, with volumes won.}
\label{tab:unseenR}
\begin{tabular}{llccc}
\toprule
reconstructor & acquisition & $6\times$ & $12\times$ & $24\times$ \\
\midrule
\multirow{3}{*}{warm CG, 1 evaluation} & protocol & 38.62 & \textbf{35.52} & \textbf{33.24} \\
 & learned & \textbf{41.08} & 35.48 & 32.45 \\
 & gain (won) & $+2.46$ (150/150) & $-0.03$ (65/150) & $-0.78$ (7/150) \\
\midrule
\multirow{3}{*}{warm CG, 3 evaluations} & protocol & 39.39 & 35.64 & 32.95 \\
 & learned & \textbf{44.94} & \textbf{38.84} & \textbf{34.25} \\
 & gain (won) & $+5.52$ (150/150) & $+3.20$ (150/150) & $+1.28$ (150/150) \\
\bottomrule
\end{tabular}
\end{table}
\subsection*{The front on all 150 volumes of the acquisition split}
\begin{table}[h]\centering\small
\caption{\textbf{The cost-quality front on the full-k-space split, all 150 validation volumes} (BraTS $60^3$, 30~dB,
best $\lambda$ per volume and rate, PSNR / SSIM). The same split and models as Tables~\ref{tab:acq} and \ref{tab:acq2x2}.
Paired margins of our reconstructor over the 20-step patch-diffusion prior: one evaluation $+2.03\,[2.0,2.1]$ /
$+3.93\,[3.9,4.0]$ / $+4.29\,[4.2,4.4]$~dB at $4/8/16\times$, three evaluations $+3.31$ / $+4.28$ / $+4.17$~dB, each on
$150/150$ volumes.}
\label{tab:front150}
\begin{tabular}{lcccc}
\toprule
method & NFE & $4\times$ & $8\times$ & $16\times$ \\
\midrule
CG-SENSE (no prior)                        & 0  & 33.26 / 0.586 & 30.60 / 0.533 & 28.94 / 0.506 \\
one-step MeanFlow prior, plug-and-play     & 4  & 36.32 / 0.652 & 32.64 / 0.573 & 30.13 / 0.528 \\
patch-diffusion prior, plug-and-play       & 20 & 38.89 / 0.805 & 33.20 / 0.701 & 30.04 / 0.659 \\
\textbf{ours}, one evaluation              & 1  & 40.91 / 0.863 & 37.12 / 0.835 & \textbf{34.33} / \textbf{0.854} \\
\textbf{ours}, three evaluations           & 3  & \textbf{42.13} / \textbf{0.895} & \textbf{37.45} / 0.832 & 34.19 / 0.799 \\
\midrule
\textbf{ours}, one evaluation, learned mask   & 1 & 45.54 / 0.934 & 39.24 / 0.896 & 33.93 / 0.815 \\
\textbf{ours}, three evaluations, learned mask & 3 & \textbf{48.86} / \textbf{0.975} & \textbf{42.66} / \textbf{0.930} & \textbf{36.68} / \textbf{0.902} \\
\bottomrule
\end{tabular}
\end{table}

\subsection*{The knee front on all 33 validation volumes}
\begin{table}[h]\centering\small
\caption{\textbf{SKM-TEA knee, all 33 validation volumes} (quarter resolution, real k-space and scanner coil maps,
30~dB, best $\lambda$ per volume and rate, PSNR / SSIM, CG-20 for our models). Paired margins over the 20-step
patch-diffusion prior: one evaluation $+0.82/+2.23/+3.07$~dB, three evaluations $+1.94/+4.34/+4.89$~dB at
$4/8/16\times$, each on $33/33$ volumes.}
\label{tab:knee33}
\begin{tabular}{lcccc}
\toprule
method & NFE & $4\times$ & $8\times$ & $16\times$ \\
\midrule
CG-SENSE (no prior)                          & 0  & 37.94 / 0.924 & 30.09 / 0.721 & 26.64 / 0.578 \\
patch-diffusion prior, plug-and-play         & 20 & 38.83 / 0.939 & 31.29 / 0.741 & 27.01 / 0.570 \\
\textbf{ours}, one evaluation                & 1  & 39.71 / 0.943 & 33.50 / 0.851 & 30.09 / 0.764 \\
\textbf{ours}, three evaluations             & 3  & \textbf{40.88} / \textbf{0.964} & \textbf{35.62} / \textbf{0.916} & \textbf{31.86} / \textbf{0.850} \\
\midrule
\textbf{ours}, one evaluation, learned mask  & 1  & 41.81 / 0.964 & 34.80 / 0.892 & 31.03 / 0.828 \\
\bottomrule
\end{tabular}
\end{table}
\section{Every evaluated run}\label{app:log}
{\scriptsize
\begin{longtable}{llrrrrrr}
\caption{Every evaluated run (best $\lambda$ per run/rate/NFE; mean$\pm$std over volumes; repaired benchmark, k-space SNR 30~dB). Source: \texttt{results/*.csv} in the released repository.}\\
\toprule
run & obj & $R$ & $\lambda^\star$ & NFE & PSNR (dB) & SSIM & s/vol \\ \midrule
\endfirsthead
\toprule
run & obj & $R$ & $\lambda^\star$ & NFE & PSNR (dB) & SSIM & s/vol \\ \midrule
\endhead
\bottomrule
\endfoot
a\_adaptive\_60\_warm & meanflow & 4 & 0.03 & 1 & 34.43$\pm$1.25 & 0.830 & 0 \\
a\_adaptive\_60\_warm & meanflow & 8 & 0.03 & 1 & 31.55$\pm$1.12 & 0.793 & 0 \\
a\_adaptive\_60\_warm & meanflow & 16 & 0.0001 & 1 & 31.06$\pm$1.15 & 0.853 & 0 \\
a\_adaptive\_60\_warm & meanflow & 32 & 0.0001 & 1 & 29.84$\pm$1.16 & 0.822 & 0 \\
\midrule
a\_fixed\_60\_std10x7\_warm & meanflow & 4 & 0.03 & 1 & 38.64$\pm$1.40 & 0.870 & 0 \\
a\_fixed\_60\_std10x7\_warm & meanflow & 8 & 0.03 & 1 & 35.01$\pm$1.32 & 0.839 & 0 \\
a\_fixed\_60\_std10x7\_warm & meanflow & 16 & 0.1 & 1 & 32.44$\pm$1.43 & 0.857 & 0 \\
a\_fixed\_60\_std10x7\_warm & meanflow & 32 & 0.1 & 1 & 31.19$\pm$1.38 & 0.838 & 0 \\
\end{longtable}}

\end{document}